\documentclass[10pt,journal,compsoc]{IEEEtran}
\usepackage{float}
\usepackage{booktabs}
\usepackage{enumitem}
\usepackage{amssymb}
\usepackage{mathtools}
\usepackage{bm}

\newcommand{\f}{\frac}
\newcommand{\p}{\partial}
\newcommand{\B}{\mathbf}
\newcommand{\V}{\mathbf{v}}
\newcommand{\M}{\mathbf{M}}
\newcommand{\name}{HOPE}

\ifCLASSOPTIONcompsoc
  \usepackage[nocompress]{cite}
\else
  \usepackage{cite}
\fi

\begin{document}
\title{\name: High-order Polynomial Expansion of Black-box Neural Networks}

\author{Tingxiong Xiao, 
        Weihang Zhang, 
        Yuxiao Cheng, 
        and Jinli Suo
\IEEEcompsocitemizethanks{\IEEEcompsocthanksitem All the authors are with the Department of Automation, Tsinghua University, Beijing, China. Jinli Suo is also affiliated with the Institute of Brain and Cognitive Sciences, Tsinghua University, and Shanghai Artificial Intelligence Laboratory, Shanghai, China. Emails: 
\{xtx22,zwh19,cyx22\}@mails.tsinghua.edu.cn; 
jlsuo@tsinghua.edu.cn.
\IEEEcompsocthanksitem Corresponding author: Jinli Suo.
}
}

\IEEEtitleabstractindextext{
\begin{abstract}
Despite their remarkable performance, deep neural networks remain mostly ``black boxes'', suggesting inexplicability and hindering their wide applications in fields requiring making rational decisions. Here we introduce \name~ (High-order Polynomial Expansion), a method for expanding a network into a high-order Taylor polynomial on a reference input. Specifically, we derive the high-order derivative rule for composite functions and extend the rule to neural networks to obtain their high-order derivatives quickly and accurately. From these derivatives, we can then derive the Taylor polynomial of the neural network, which provides an explicit expression of the network's local interpretations. Numerical analysis confirms the high accuracy, low computational complexity, and good convergence of the proposed method. Moreover, we demonstrate \name's wide applications built on deep learning, including function discovery, fast inference, and feature selection. The code is available at https://github.com/HarryPotterXTX/HOPE.git.
\end{abstract}

}

\maketitle
\IEEEdisplaynontitleabstractindextext
\IEEEpeerreviewmaketitle

\IEEEraisesectionheading{\section{Introduction}\label{sec:introduction}}
\IEEEPARstart{D}{eep} neural networks have gained widespread adoption due to their ability for universal approximation, as proved by numerous studies \cite{chen1995universal,cybenko1989approximation,hornik1989multilayer,
leshno1993multilayer}. However, deep learning networks are largely considered black boxes that hinder their practical applications. Therefore, understanding the rationale behind predictions is crucial when making relatively logical decisions based on the network output or deciding whether to deploy a new model. 
This requirement for understanding is particularly important in areas such as clinical decision-making \cite{caruana2015intelligible}\cite{antoniadi2021current}, drug discovery \cite{jimenez2020drug}, and physical law identification \cite{xie2022data}, so people often prioritize models that align with their intuition over accuracy \cite{daneman2015blood,rankovic2017cns,leeson2015molecular}.
Therefore, there is a growing need for explainable AI (XAI) approaches to make deep learning more transparent and convincing \cite{goebel2018explainable,murdoch2019definitions,zachary2018mythos}.

XAI can be divided into five categories\cite{jimenez2020drug}: feature attribution, instance-based, graph-convolution-based, self-explaining, and uncertainty estimation, among which the feature attribution family calculates the relevance of every input feature for the final prediction and has been the most widely used XAI in recent years. The implementations identifying the feature attribution can be further grouped into the following three categories.

{\em Gradient-based feature attribution} approaches measure the impact of a change within an input neighborhood on the change in the output. These methods are mainly inspired by back-propagation\cite{rumelhart1986learning}, and some well-known methods include saliency map\cite{simonyan2013deep}, Integrated Gradients\cite{sundararajan2017axiomatic}, SmoothGrad\cite{smilkov2017smoothgrad}, local explanation vectors\cite{baehrens2010explain}, Grad-CAM\cite{selvaraju2017grad}, guided backpropagation\cite{springenberg2014striving}, LRP\cite{bach2015pixel}, and deep Taylor decomposition\cite{montavon2017explaining}. 
The gradient-based methods calculate the first-order derivative of the model as the feature attribution, but omit the higher-order terms, which are important for nonlinear functions. 

{\em Surrogate-model feature attribution} aims at developing a surrogate explanatory model to the original function, to mirror the computational logic of the original model. The representative surrogate models include LIME\cite{ribeiro2016should}, DeepLIFT\cite{shrikumar2017learning}, Shapley value\cite{strumbelj2010efficient}, LRP\cite{bach2015pixel}, SHAP\cite{lundberg2017unified}, and BETA\cite{lakkaraju2017interpretable}. However, such surrogate models are mostly linear and suffer from insufficient  approximation accuracy. 

{\em Perturbation-based feature attribution} modifies or removes part of the input to measure the corresponding change in the model output, which reflects the feature importance of the neural network input. Methods like feature masking\cite{vstrumbelj2009explaining}, perturbation analysis\cite{fong2017interpretable}, response randomization\cite{olden2002illuminating}, and conditional multivariate models\cite{zintgraf2017visualizing} fall into this category. Although intuitive, perturbation-based methods are computationally slow when the number of input features increases\cite{zintgraf2017visualizing}, and the final result tends to be strongly influenced by the number of perturbed features \cite{ancona2017towards}.

In summary, an approach capable of approximating a general deep neural network with high accuracy, low cost and good interpretability is demanded. Here we present \name~ (High-order Polynomial Expansion), an approach to expand a deep neural network into a high-order Taylor polynomial on a reference input. The Taylor expansion is built on calculating the derivatives of the target neural network, which is intrinsically a nonlinear function.  
We first derive the high-order derivative rule for composite functions and extend this to neural networks to obtain the high-order derivatives quickly and accurately.
Our method serves as a gradient-based method and a surrogate model, as it integrates the high-order derivatives and uses nonlinear polynomials to locally approximate the neural network. 
Our expansion is of high approximation accuracy, and our computational cost is far lower than perturbation-based methods because we can get all derivatives with only one back-propagation.

Actually, in recent years, some researchers have paid attention to the Taylor expansion of neural networks. LRP\cite{bach2015pixel} and deep Taylor decomposition\cite{montavon2017explaining} provide a first-order approximation, but they neglect the high-order terms.
Morala P et al. \cite{morala2021towards} explored a mathematical framework for expanding a single-hidden-layer neural network, but it does not apply to multi-layer networks. NN-Poly\cite{zhu2022nn}, on the other hand, can infer the Taylor expansion of single-layer neural networks, and obtain the Taylor polynomial of the multi-layer neural network through forward composition, but at extremely high computational complexity.
SHoP\cite{xiao2023shop} proposed a Taylor expansion framework to solve high-order partial differential equations, but it only considered fully connected neural networks and adopted a layer-wise approach for the derivation process, rather than a module-wise approach. 
Differently, our method is similar to back-propagation and computes the derivative of the final output with respect to the intermediate output, propagating from the output layer back to the input layer. Compared to forward composition, our method, \name, has significant advantages in terms of accuracy, speed, computational complexity, and memory consumption.

\vspace{2mm}
To summarize, the main technical contributions are listed as follows:
\begin{itemize}[leftmargin=*,itemsep=0pt,topsep=2pt,parsep=0pt]
\item We infer the high-order derivative rule 
and extend it for calculating the derivatives of deep neural networks with higher accuracy, higher speed, and less memory consumption than the conventional computational-graph-based counterpart.
\item We propose a framework to expand a deep neural network into a Taylor series, providing an explicit explanation of the inner workings of the ``black-box'' model.
\item We prove the equivalence between a neural network and its Taylor series, and analyze its convergence condition. 
\item We demonstrate the wide applications of Taylor expanded deep neural networks, such as function discovery, fast inference, and feature selection.  
\end{itemize}

\vspace{2mm}
\noindent The paper is structured as follows: In Sec.~\ref{High-order Derivative Rule for Composite Functions}, we present the high-order derivative rule for composite functions. In Sec.~\ref{High-order Derivative Rule for Neural Networks}, we extend this rule to deep neural networks, which allows for efficient and accurate computation of their high-order derivatives. Based on the calculated derivatives, we propose the Taylor expansion framework---\name, and analyze its bounds,  convergence, and computational complexity in Sec. \ref{Taylor Expansion for Neural Networks}. 
In Sec.~\ref{Experiments}, we conducted a series of experiments to show \name's significant advantages over the computational-graph-based method, verified the convergence condition, and demonstrated its applications in function discovery, fast inference, and feature selection. Finally, in Sec.\ref{Conclusion}, we analyze the advantages and disadvantages of the proposed approach and discuss some possible future work.

\section{High-order Derivative Rule for Composite Functions}
\label{High-order Derivative Rule for Composite Functions}

Considering a composite function 
\begin{equation}
    \B{y}=h(g(\B{x})),
\end{equation} 
which is constructed by two functions $\B{z}=g(\B{x})$ and $\B{y}=h(\B{z})$ 
to map the input $\B{x}\in \mathbb{R}^p$ to the intermediate state variable $\B{z}\in \mathbb{R}^s$, and then to the final output $\B{y}\in \mathbb{R}^o$ sequentially. Assuming the two functions $g(\cdot)$ and $f(\cdot)$ are $n$-order differentiable at $\B x0$ and $\B z0=g(\B x0)$ respectively, this section will derive the high-order derivative rule for the composite function $\B{y}=h(g(\B{x}))$ in three systems: Single-Input Single-State Single-Output (SISSSO), Multiple-Input Multiple-State Single-Output (MIMSSO), and Multiple-Input Multiple-State Multiple-Output (MIMSMO).

\subsection{SISSSO}

For an SISSSO system, $\B x,\B{z},\B{y}\in \mathbb{R}$. From the chain rule, the first three derivatives of $\B{y}=h(g(\B{x}))$ can be calculated as
\begin{equation}  
\begin{split}
\left\{  
\begin{array}{lc}
    \f{\p\B y}{\p\B x}
    =\f{\p\B z}{\p\B x}\f{\p\B y}{\p\B z},  \\    
    \f{{\p}^2\B y}{\p\B x^2}
    =\f{\p^2\B z}{\p\B x^2}\f{\p\B y}{\p\B z}+(\f{\p\B z}{\p\B x})^2\f{\p^2\B y}{\p\B z^2}, \\
    \f{{\p}^3\B y}{\p\B x^3}
    =\f{\p^3\B z}{\p\B x^3}\f{\p\B y}{\p\B z}+3\f{\p\B z}{\p\B x}\f{\p^2\B z}{\p\B x^2}\f{\p^2\B y}{\p\B z^2}+(\f{\p\B z}{\p\B x})^3\f{\p^3\B y}{\p\B z^3}.
\end{array}
\right.  
\label{3-order derivatives}
\end{split}
\end{equation} 
For more terms, we can convert $\f{{\p}^k\B y}{\p\B z^{k-1} \p\B x}$ to $\f{\p\B z}{\p\B x}\f{{\p}^k\B y}{\p\B z^k}$, and calculate $\f{{\p}^n\B y}{\p\B x^n}$ from $\{\f{\p^k\B y}{\p\B z^k}, k=1,\ldots,n\}$ and $\{\f{\p^k\B z}{\p\B x^k}, k=1,\ldots,n\}$. Hence, Eq.~(\ref{3-order derivatives}) turns into following matrix form
\begin{equation}
\begin{split}
\left[
    \begin{array}{c}
            \f{\p\B y}{\p\B x}  \\
            \vdots \\
            \f{\p^n\B y}{\p\B x^n}
    \end{array}
\right]=
\left[
    \begin{array}{cccc}
            \f{\p\B \B z}{\p\B \B x} & 0 & 0 & 0  \\
            \f{\p^2\B z}{\p\B x^2} & (\f{\p\B z}{\p\B x})^2 & 0 & 0 \\
            \f{\p^3\B z}{\p\B x^3} & 3\f{\p\B z}{\p\B x}\f{\p^2\B z}{\p\B x^2} & (\f{\p\B z}{\p\B x})^3 & 0 \\
            \vdots & \vdots & \vdots & \ddots
    \end{array}
\right]
\left[
    \begin{array}{c}
            \f{\p\B y}{\p\B z}  \\
            \vdots \\
            \f{\p^n\B y}{\p\B z^n}
    \end{array}
\right],
\end{split}
\label{chain matrix}
\end{equation}
which can be further abbreviated as
\begin{equation}
    \V^{y,x}=\M^{z,x}\V^{y,z}.
\label{abbreviated chain matrix}
\end{equation}
In this equation $\V^{y,x}, \V^{y,z} \in \mathbb{R}^n$ are respectively the vectors composed of derivatives $\{\f{\p^k\B y}{\p\B x^k}\}$ and $\{\f{\p^k\B y}{\p\B z^k}\}$; $\M^{z,x} \in \mathbb{R}^{n\times n}$ is the transformation matrix composed of 
$\f{\p^k\B z}{\p\B x^k}$ and takes a lower triangular form. So far, the calculation of $h(g(\B x))$'s $n$-order derivatives turns into the computation of $\M^{z,x}$. 

From Eq. (\ref{chain matrix}) the $i$th and $i+1$th terms ($i<n$) are respectively
\begin{equation}
    \f{\p^i \B y}{\p \B x^i}=\sum_{j=1}^{n}\M^{z,x}_{i,j}\f{\p^j \B y}{\p \B z^j}
    \label{eq:ith}
\end{equation}
and
\begin{equation}
    \f{\p^{i+1} \B y}{\p \B x^{i+1}}=\sum_{j=1}^{n}\M^{z,x}_{i+1,j}\f{\p^j \B y}{\p \B z^j}.
    \label{eq:i+1th}
\end{equation}
Taking derivatives over both sides of Eq.~(\ref{eq:ith}) we arrive at
\begin{equation}
\begin{split}
    \f{\p^{i+1}\B y}{\p\B x^{i+1}}
    &=\sum_{j=1}^{n}\f{\p }{\p\B x}(\M^{z,x}_{i,j}\f{\p^j\B y}{\p\B z^j}) \\
    &=\sum_{j=1}^{n}\f{\p }{\p\B x}\M^{z,x}_{i,j}\f{\p^j\B y}{\p\B z^j}+\sum_{j=1}^{n}\f{\p\B z}{\p\B x}\M^{z,x}_{i,j}\f{\p^{j+1}\B y}{\p\B z^{j+1}} \\
    &=\sum_{j=1}^{n}\left(\f{\p }{\p\B x}\M^{z,x}_{i,j}
    +\f{\p\B z}{\p\B x}\M^{z,x}_{i,j-1} \right)\f{\p^j\B y}{\p\B z^j} \\
    &\quad-\f{\p\B z}{\p\B x}\M^{z,x}_{i,0}\f{\p\B y}{\p\B z}
    +\f{\p\B z}{\p\B x}\M^{z,x}_{i,n}\f{\p^{n+1}\B y}{\p\B z^{n+1}}.
\end{split}
\label{eq:i+1th(1)}
\end{equation}
Because $\M^{z,x}_{i,0}=0$ and $\M^{z,x}_{i,n}=0~(i < n)$, Eq.~(\ref{eq:i+1th(1)}) can be simplified into
\begin{equation}
    \f{\p^{i+1}\B y}{\p\B x^{i+1}}
    =\sum_{j=1}^{n}\left(\f{\p }{\p\B x}\M^{z,x}_{i,j}
    +\f{\p\B z}{\p\B x}\M^{z,x}_{i,j-1} \right)\f{\p^j\B y}{\p\B z^j}.
    \label{eq:i+1th(2)}
\end{equation}
Compare Eq.~(\ref{eq:i+1th}) and Eq.~(\ref{eq:i+1th(2)}), we can get the recurrence formula of  $\M^{z,x}$ as
\begin{equation}  
\left\{  
    \begin{array}{lc}
        \M^{z,x}_{1,1}=\f{\p z}{\p x}, &  \\  
        \M^{z,x}_{i,j}=0, & i < j \\ 
        \M^{z,x}_{i,j}=\f{\p \M^{z,x}_{i-1,j}}{\p x}
         +\f{\p z}{\p x}\M^{z,x}_{i-1,j-1}, & i\geq j
    \end{array}  
\right.
\label{recurrence}
\end{equation} 
which explicitly composes the $n$-order transformation matrix $\M^{z,x}$ in Eq.~(\ref{abbreviated chain matrix}).

\subsection{MIMSSO}

\vspace{2mm}
\noindent\textbf{Unmixed partial derivatives.\quad}
For a MIMSSO system with $p$-dimensional input and $s$ intermediate states, i.e., $\B x\in \mathbb{R}^p$, $\B z\in \mathbb{R}^s$, $\B y\in \mathbb{R}$, we can get the first three unmixed partial derivatives of $\B{y}=h(g(\B{x}))$ as
\begin{equation}  
\begin{split}
\left\{  
\begin{array}{lc}
    \f{\p\B y}{\p\B x_i}
    =\sum_{j=1}^s\left(\f{\p\B z_j}{\p\B x_i}\f{\p\B y}{\p\B z_j}\right),  \\    
    \f{{\p}^2\B y}{\p\B x_i^2}
    =\sum_{j=1}^s\left(\f{\p^2\B z_j}{\p\B x_i^2}\f{\p\B y}{\p\B z_j}+(\f{\p\B z_j}{\p\B x_i})^2\f{\p^2\B y}{\p\B z_j^2}\right), \\
    \f{{\p}^3\B y}{\p\B x_i^3}
    =\sum_{j=1}^s\left(\f{\p^3\B z_j}{\p\B x_i^3}\f{\p\B y}{\p\B z_j}+3\f{\p\B z_j}{\p\B x_i}\f{\p^2\B z_j}{\p\B x_i^2}\f{\p^2\B y}{\p\B z_j^2}+(\f{\p\B z_j}{\p\B x_i})^3\f{\p^3\B y}{\p\B z_j^3}\right).
\end{array}
\right.  
\label{3-order derivatives(2)}
\end{split}
\end{equation} 
To facilitate derivation, we define an operator $\beta$ to save the information of their $k$-order unmixed partial derivatives
\begin{equation}
\begin{split}
\f{\beta^k }{\beta \B x^k}
=\left[
    \begin{array}{c}
        \f{\p^k }{\p\B x_1^k} \\
        \vdots \\
        \f{\p^k }{\p\B x_p^k}
    \end{array}
\right],
\label{beta definition}
\end{split}
\end{equation}
and the following equations hold 
\begin{equation}
\begin{split}
\f{\beta^k \B z^T}{\beta \B x^k}
&=\left[
    \begin{array}{ccc}
        \f{\p^k\B z_1}{\p\B x_1^k} & \ldots & \f{\p^k\B z_s}{\p\B x_1^k} \\
        \vdots & \ddots & \vdots \\
        \f{\p^k\B z_1}{\p\B x_p^k} & \ldots & \f{\p^k\B z_s}{\p\B x_p^k}
    \end{array}
\right],\\
\f{\beta^k \B y}{\beta \B x^k} 
&=\left[
    \begin{array}{c}
        \f{\p^k\B y}{\p\B x_1^k} \\
        \vdots \\
        \f{\p^k\B y}{\p\B x_p^k}
    \end{array}
\right]~\text{and}~
\f{\beta^k \B y}{\beta \B z^k}
=\left[
    \begin{array}{c}
        \f{\p^k\B y}{\p\B z_1^k} \\
        \vdots \\
        \f{\p^k\B y}{\p\B z_s^k}
    \end{array}
\right].
\label{beta use definition}
\end{split}
\end{equation}
Based on the above definitions, Eq.~(\ref{3-order derivatives(2)}) can be rewritten as
\begin{equation}  
\begin{split}
\left\{  
\begin{array}{lc}
    \f{\beta\B y}{\beta\B x}
    =\f{\beta \B z^T}{\beta \B x}\f{\beta\B y}{\beta\B z},  \\    
    \f{\beta^2\B y}{\beta\B x^2}
    =\f{\beta^2\B z^T}{\beta\B x^2}\f{\beta\B y}{\beta\B z}
    +{(\f{\beta\B z^T}{\beta\B x})}^{\circ 2}\f{\beta^2\B y}{\beta\B z^2}, \\
    \f{\beta^3\B y}{\beta\B x^3}
    =\f{\beta^3\B z^T}{\beta\B x^3}\f{\beta\B y}{\beta\B z}
    +(3\f{\beta\B z^T}{\beta\B x}\odot\f{\beta^2\B z^T}{\beta\B x^2})\f{\beta^2\B y}{\beta\B z^2}
    +{(\f{\beta\B z^T}{\beta\B x})}^{\circ 3}\f{\beta^3\B y}{\beta\B z^3},
\end{array}
\right.  
\label{3-order derivatives beta}
\end{split}
\end{equation} 
where $\circ k$ is Hadamard power, $(\mathbf{A}^{\circ k})_{i,j}={(\mathbf{A}_{i,j})}^k$, and $\odot$ is Hadamard product, $(\mathbf{A}\odot\mathbf{B})_{i,j}=\mathbf{A}_{i,j}\mathbf{B}_{i,j}$. 

Similar to Eq.~(\ref{chain matrix}), we turn Eq.~(\ref{3-order derivatives beta}) into a matrix form
\begin{small} 
\begin{equation}
\begin{split}
\left[\!
    \begin{array}{c}
            \f{\beta\B y}{\beta\B x}  \\
            \vdots \\
            \f{\beta^n\B y}{\beta\B x^n}
    \end{array}
\!\right]\!=\!
\left[\!
    \begin{array}{cccc}
            \f{\beta \B z^T}{\beta \B x} & 0 & 0 & 0  \\
            \f{\beta^2\B z^T}{\beta\B x^2} & {(\f{\beta\B z^T}{\beta\B x})}^{\circ 2} & 0 & 0 \\
            \f{\beta^3\B z^T}{\beta\B x^3} & 3\f{\beta\B z^T}{\beta\B x}\odot\f{\beta^2\B z^T}{\beta\B x^2} & {(\f{\beta\B z^T}{\beta\B x})}^{\circ 3} & 0 \\
            \vdots & \vdots & \vdots & \ddots
    \end{array}
\!\right]
\left[\!
    \begin{array}{c}
            \f{\beta\B y}{\beta\B z}  \\
            \vdots \\
            \f{\beta^n\B y}{\beta\B z^n}
    \end{array}
\!\right],
\end{split}
\label{chain matrix beta}
\end{equation}
\end{small}
which is of a consistent form with Eq.~(\ref{chain matrix}), only with scalar elements replaced by matrices, and the power and multiplication turn into Hadamard power $\circ k$ and Hadamard product $\odot$ respectively. We further abbreviated the above equation as
\begin{equation}
    \V^{y,x}=\M^{z,x}\V^{y,z}.
\label{abbreviated chain matrix beta}
\end{equation}

\vspace{2mm}
\noindent\textbf{Mixed partial derivatives.\quad}
The first module of the neural network is mostly a linear layer, such as a fully connected layer or convolutional layer that satisfies 
\begin{equation}  
\f{\p^k\B z_j}{\p\B x_{i_1}\ldots\p\B x_{i_k}}=0~(k>1).
\label{mixed assumption}
\end{equation} 
The first three mixed derivatives of $\B{y}=h(g(\B{x}))$ are calculated as
\begin{equation}  
\begin{split}
\left\{  
\begin{array}{lc}
    \f{\p\B y}{\p\B x_{i_1}}
    =\sum_{j=1}^s\f{\p\B z_j}{\p\B x_{i_1}}\f{\p\B y}{\p\B z_j},  \\    
    \f{{\p}^2\B y}{\p\B x_{i_1}\p\B x_{i_2}}
    =\sum_{j=1}^s\f{\p\B z_j}{\p\B x_{i_1}}\f{\p\B z_j}{\p\B x_{i_2}}\f{\p^2\B y}{\p\B z_j^2}, \\
    \f{{\p}^3\B y}{\p\B x_{i_1}\p\B x_{i_2}\p\B x_{i_3}}
    =\sum_{j=1}^s\f{\p\B z_j}{\p\B x_{i_1}}\f{\p\B z_j}{\p\B x_{i_2}}\f{\p\B z_j}{\p\B x_{i_3}}\f{\p^3\B y}{\p\B z_j^3}.
\end{array}
\right.  
\label{3-order derivatives(4)}
\end{split}
\end{equation} 
Based on the definition in Eqns.~(\ref{beta definition})(\ref{beta use definition}), the above equations turns into 
\begin{equation}  
\begin{split}
\left\{  
\begin{array}{lc}
    \f{\beta\B y}{\beta\B x}
    =\B Q_1\f{\beta\B y}{\beta\B z},  \\    
    \f{\beta}{\beta\B x}\otimes \f{\beta\B y}{\beta\B x}
    =\B Q_2\f{\beta^2\B y}{\beta\B z^2}, \\
    \f{\beta}{\beta\B x}\otimes\f{\beta}{\beta\B x}\otimes\f{\beta\B y}{\beta\B x}
    =\B Q_3\f{\beta^3\B y}{\beta\B z^3}, \\
    \B Q_1=\f{\beta \B z^T}{\beta \B x}, \B Q_k=\left(\f{\beta \B z^T}{\beta \B x}\otimes\B 1^{p^{k-1}}\right)\odot\left(\B 1^p\otimes \B Q_{k-1}\right),
\end{array}
\right.  
\label{3-order derivatives beta mixed}
\end{split}
\end{equation} 
where $\otimes$ is Kronecker product, and $\B 1^p\in\mathbb{R}^p$ is an all-1 column vector. Similar to Eqns.~(\ref{chain matrix})(\ref{chain matrix beta}), we rewrite Eq.~(\ref{3-order derivatives beta mixed}) into the matrix form
\begin{equation}
\begin{split}
\left[
    \begin{array}{c}
            \f{\beta\B y}{\beta\B x}  \\
            \vdots \\
            \left(\f{\beta}{\beta\B x}\otimes\right)^{n-1}\f{\beta\B y}{\beta\B x}
    \end{array}
\right]=
\left[
    \begin{array}{cccc}
            \B Q_1 &  & 0 \\
             & \ddots &  \\
             0 &  & \B Q_n
    \end{array}
\right]
\left[
    \begin{array}{c}
            \f{\beta\B y}{\beta\B z}  \\
            \vdots \\
            \f{\beta^n\B y}{\beta\B z^n}
    \end{array}
\right],
\end{split}
\label{chain matrix beta mixed}
\end{equation}
with
\begin{small}
\begin{equation}
\begin{split}
    \left(\f{\beta}{\beta\B x}\otimes\right)^{k-1}\f{\beta\B y}{\beta\B x}
    =\underbrace{\f{\beta}{\beta\B x}\otimes\ldots\f{\beta}{\beta\B x}\otimes}_{k-1} \f{\beta\B y}{\beta\B x}
    =\left[
    \begin{array}{c}
        \f{\p^k\B y}{\p\B x_1^k} \\
        \f{\p^k\B y}{\p\B x_1^{k-1}\p\B x_2} \\
        \vdots \\
        \f{\p^k\B y}{\p\B x_p^{k-1}\p\B x_{p-1}} \\
        \f{\p^k\B y}{\p\B x_p^k}
    \end{array}
\right],
\end{split}
\end{equation}
\end{small} 
which contains all the $k$-order partial derivatives. Note that this formula is derived with a limitation of Eq.~(\ref{mixed assumption}), which means only linear modules can apply it.

We further abbreviate Eq.~(\ref{chain matrix beta mixed}) as
\begin{equation}
    {\V^\star}^{y,x}={\M^\star}^{z,x}\V^{y,z}.
\label{abbreviated chain matrix beta mixed}
\end{equation}

\subsection{MIMSMO}
For a MIMSMO system with $\B x\in \mathbb{R}^p$, $\B z\in \mathbb{R}^s$, $\B y\in \mathbb{R}^o$, we only need to apply the same formula for derivative calculation of a MIMSSO to each output entry, and the unmixed formula and mixed formula is
\begin{equation}  
\begin{split}
\left\{  
\begin{array}{lc}
    \V^{y_i,x}=\M^{z,x}\V^{y_i,z}, & i=1,\ldots,o\\
    {\V^\star}^{y_i,x}={\M^\star}^{z,x}\V^{y_i,z}. & i=1,\ldots,o
\end{array}
\right.  
\label{MIMSMO}
\end{split}
\end{equation} 

\section{High-order Derivative Rule for Neural Networks}
\label{High-order Derivative Rule for Neural Networks}

\begin{figure*}[!t]
\centering
\centerline{\includegraphics[width=\textwidth]{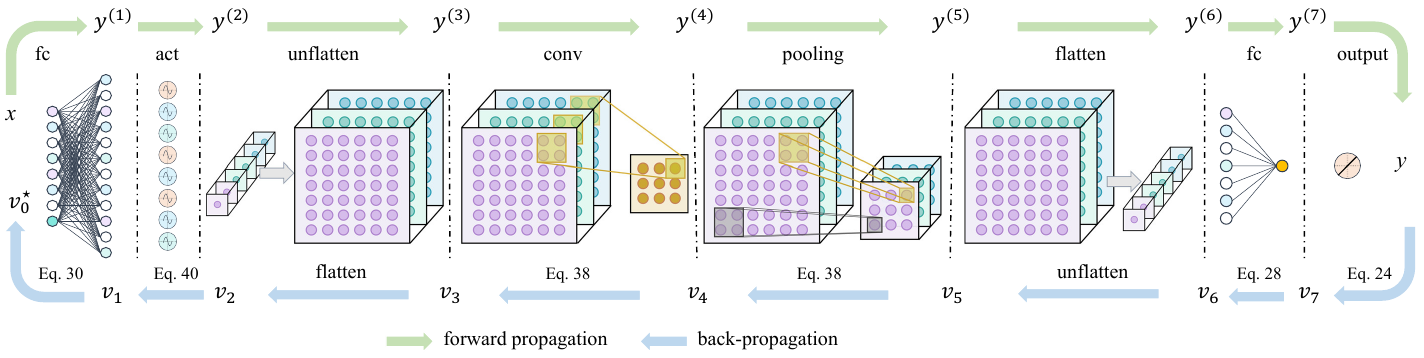}}
\hfil
\vspace{-2mm}
\caption{\textbf{The framework for calculating the high-order derivatives of \name.} The network has seven modules: fully connected layer (fc), activation function (act), unflatten module (unflatten), convolutional layer (conv), pooling layer (pooling), flatten module (flatten), and single-output fully connected layer. The output of $m$th module is $\B y^{(m)}$ and the final output $\B y=\B y^{(7)}$. \name~ is similar to back-propagation in that we calculate the derivatives of $\B y$ with respect to the intermediate output $\B y^{(m)}$, starting from the output layer and moving backward towards the input layer. In the intermediate layer, we only need to calculate $\B v_m$, which contains all the unmixed partial derivatives. In the input layer, we can calculate $\B v^\star_m$, which contains all the mixed partial derivatives.}
\label{propagation}
\vspace{-3mm}
\end{figure*}

This section mainly introduces the high-order derivative rule of a deep neural network. Since a multiple-output network can be regarded as multiple single-output networks, we consider only the single-output cases. 
Without loss of generality, we derive the back-propagation of high-order derivatives of the most common modules, with the network structure illustrated in Fig.~\ref{propagation}. 

Before proceeding with the detailed derivations, we define the following notations. The input is denoted as $\B x$, the output of $m$th module as $\B y^{(m)}$, the length of $\B y^{(m)}$ as $o_m$, and the final output as $\B y$. To simplify the expression, we omit the superscripts of $\B v^{\B y,\B y^{(m)}}$ and $\B M^{\B y^{(m)},\B y^{(m-1)}}$ respectively as $\B v_m$ and $\B M_m$.

\subsection{Output Unit}

As for the final output $\B y=\B y^{(7)}\in \mathbb{R}$, according to the definition in Eq.~(\ref{beta use definition}), its $k$-order derivatives can be calculated as
\begin{equation}
\begin{split}
    \f{\beta^k \B y}{\beta {\B y^{(7)}}^k}
    =\left[
    \begin{array}{c}
        \f{\p^k\B y}{\p{\B y^{(7)}}^k}
    \end{array}
    \right]
    =\left\{
    \begin{array}{cc}
    \left[1\right], & k=1 \\
    \left[0\right], & k>1.
    \end{array}
    \right.
\end{split}
\end{equation}
Further from Eq.~(\ref{abbreviated chain matrix beta}), we can obtain the  initial derivative in vector form
\begin{equation}
\begin{split}
    \B v_7
    =\left[
    \begin{array}{ccc}
        {\f{\beta \B y}{\beta {\B y^{(7)}}}}^T &
        \cdots &
        {\f{\beta^n \B y}{\beta {\B y^{(7)}}^n}}^T
    \end{array}
    \right]^T
    =\left[
    \begin{array}{cccc}
        1 &
        0 &
        \cdots &
        0
    \end{array}
    \right]^T.
\end{split}
\end{equation}

\subsection{Fully Connected Layer}
For a fully connected layer, its input-output relationship is defined as
\begin{equation}
\begin{split}
    \B y^{(m+1)}=\B W^{(m+1)}\B y^{(m)}+\B b^{(m+1)},
\end{split}
\end{equation}
where $\B W^{(m+1)}\in \mathbb{R}^{o_{m+1}\times o_{m}}$ is the weight matrix, and $\B b^{(m+1)} \in \mathbb{R}^{o_{m+1}}$ is the bias vector. The $k$-order derivative of the $i$-th node of $\B y^{(m+1)}$ with respect to the $j$-th node of $\B y^{(m)}$ is 
\begin{equation}
\begin{split}
    \f{\p^k\B y^{(m+1)}_i}{\p {\B y^{(m)}}_j^k}
    =\left\{
    \begin{array}{cc}
    \B W^{(m+1)}_{i,j}, & k=1 \\
    0, & k>1.
    \end{array}
    \right.
\end{split}
\end{equation}
Combining with the definition in Eq.~(\ref{beta use definition}), we have
\begin{equation}
\begin{split}
\f{\beta^k {\B y^{(m+1)}}^T}{\beta {\B y^{(m)}}^k}
=\left\{
    \begin{array}{cc}
    \B W^T, & k=1 \\
    \B 0, & k>1.
    \end{array}
\right. 
\end{split}
\end{equation}

On the one hand, we can get all the unmixed partial derivatives $\{\f{\p^k\B y}{\p{\B y^{(m)}_i}^k},i=1,\ldots,o_m\}$ by calculating $\B v_m$
\begin{equation}
\begin{split}
    \B v_m=\B M_{m+1}\B v_{m+1},
\end{split}
\label{fc vector}
\end{equation}
with the transformation matrix $\B M_{m+1}$ in Eq.~(\ref{chain matrix beta}) rewritten as a block diagonal matrix
\begin{equation}
\begin{split}
\B M_{m+1}=diag(\B W^T,{\B W^T}^{\circ 2},\ldots,{\B W^T}^{\circ n}).
\label{diag unmixed}
\end{split}
\end{equation}
On the other hand, we can obtain all the mixed partial derivatives $\{\f{\p^k\B y}{\p\B y^{(m)}_{i_1}\ldots\p\B y^{(m)}_{i_k}},i_1,\ldots,i_k=1,\ldots,o_m\}$ by calculating $\B v^\star_m$ from
\begin{equation}
\begin{split}
    \B v^\star_m=\B {M^\star}_{m+1}\B v_{m+1}.
    \label{vstar}
\end{split}
\end{equation}
In this equation, the transformation matrix $\B {M^\star}_{m+1}$ defined in Eq.~(\ref{chain matrix beta mixed}) can be also rewritten as a block diagonal matrix
\begin{equation}
\begin{split}
\B {M^\star}_{m+1}=diag(\B Q_1,\B Q_2,\ldots,\B Q_n),
\label{diag mixed}
\end{split}
\end{equation}
with $\B Q_1=\B W^T$, $\B Q_k=\left(\B W^T\otimes\B 1^{({{o_m}^{k-1}})}\right)\odot\left(\B 1^{o_m}\otimes \B Q_{k-1}\right)$. 

\subsection{Convolutional Layer}
The input-output relationship of a convolutional layer can be described as 
\begin{equation}
\begin{split}
    \B y^{(m+1)}=\B y^{(m)}*\B F^{(m+1)},
\end{split}
\end{equation}
where $\B F^{(m+1)}$ is a convolutional kernel, and $*$ denotes convolutional operation. Although a convolutional layer can be regarded as a fully connected layer with a sparse weight matrix and zero bias,
taking derivative is both time consuming and memory demanding if transformed into a fully connected layer. Therefore, we derive the high-order derivative rule on the convolution representation.

The $u$-th output sums over the product of some elements in $\B y^{(m)}$ and all elements in $\B F^{(m+1)}$,
\begin{equation}
\begin{split}
    \B y^{(m+1)}_u=\sum_{v} \B y^{(m)}_v\B F^{(m+1)}_{u,v},
\end{split}
\end{equation}
where $\B F^{(m+1)}_{u,v}$ is the weight between $\B y^{(m+1)}_u$ and $\B y^{(m)}_v$. We can calculate the derivatives as
\begin{equation}
\begin{split}
    \f{\p \B y}{\p \B y^{(m)}_v}=\sum_{u} \f{\p \B y}{\p \B y^{(m+1)}_u}\f{\p \B y^{(m+1)}_u}{\p \B y^{(m)}_v}=\sum_{u} \f{\p \B y}{\p \B y^{(m+1)}_u}\B F^{(m+1)}_{u,v}.
    \label{1-order}
\end{split}
\end{equation}
Because $\f{\p {\B y^{(m+1)}_u}^k}{\p {\B y^{(m)}_v}^k}=0~(k>1)$, the high-order derivatives are
\begin{small}
\begin{equation}
\begin{split}
    \f{\p^k \B y}{\p {\B y^{(m)}_v}^k}=\sum_{u} \f{\p^k \B y}{\p {\B y^{(m+1)}_u}^k}\left({\f{\p \B y^{(m+1)}_u}{\p \B y^{(m)}_v}}\right)^k=\sum_{u} \f{\p^k \B y}{\p {\B y^{(m+1)}_u}^k}{\B F^{(m+1)}_{u,v}}^k.
    \label{k-order}
\end{split}
\end{equation}
\end{small}
Taking the matrix form, Eq.~(\ref{1-order}) turn into 
\begin{equation}
\begin{split}
    \f{\beta \B y}{\beta \B y^{(m)}}=\f{\beta \B y}{\beta \B y^{(m+1)}}*rot180(\B F).
\end{split}
\end{equation}
Comparing Eq.~(\ref{1-order}) and Eq.~(\ref{k-order}), we can also get the matrix form of the high-order derivatives as
\begin{equation}
\begin{split}
    \f{\beta^k \B y}{\beta {\B y^{(m)}}^k}=\f{\beta^k \B y}{\beta {\B y^{(m+1)}}^k}*rot180(\B F^{\circ k}).
    \label{conv k-order}
\end{split}
\end{equation}

Since the input of a convolution layer is usually image-like data, we only calculate unmixed partial derivatives as
\begin{equation}
\begin{split}
    \B v_m=\left[
    \begin{array}{ccc}
        \left(\f{\beta \B y}{\beta \B y^{(m)}}\right)^T & \ldots & \left(\f{\beta^n \B y}{\beta {\B y^{(m)}}^n}\right)^T
    \end{array}
    \right]^T
\end{split}
\end{equation}
One can also calculate the mixed partial derivatives by converting the convolutional layer into a fully connected counterpart and then applying Eq.~(\ref{vstar}) directly.

\subsection{Nonlinear Activation Layer}
Consider a nonlinear activation layer $\B y^{(m+1)}=\sigma (\B y^{(m)})$, where $\B y^{(m+1)}_i=\sigma (\B y^{(m)}_i)$, we have
\begin{equation}
\begin{split}
\f{\p^k\B y^{(m+1)}_i}{\p{\B y^{(m)}_j}^k}
=\left\{
\begin{array}{cc}
    \sigma^{(k)} (\B y^{(m)}_j), & i=j \\
    0, & i \neq j
\end{array}
\right.
\end{split}
\end{equation}
where $\sigma^{(k)} (\cdot)$ is the $k$-order derivative of this activation function. According to the definition in Eq.~(\ref{beta use definition}), we can get 

\begin{small}
\begin{equation}
\begin{split}
\f{\beta^k {\B y^{(m+1)}}^T}{\beta {\B y^{(m)}}^k}
=\left[
    \begin{array}{ccc}
        \f{\p^k\B y^{(m+1)}_1}{\p{\B y^{(m)}_1}^k} & \ldots & \f{\p^k\B y^{(m+1)}_{o_{m+1}}}{\p{\B y^{(m)}_1}^k} \\
        \vdots & \ddots & \vdots \\
        \f{\p^k\B y^{(m+1)}_1}{\p{\B y^{(m)}_{o_{m}}}^k} & \ldots & \f{\p^k\B y^{(m+1)}_{o_{m+1}}}{\p{\B y^{(m)}_{o_{m}}}^k}
    \end{array}
\right] \\
=diag\left(\sigma^{(k)} (\B y^{(m)}_1), \ldots, \sigma^{(k)} (\B y^{(m)}_{o_m})\right).
\end{split}
\end{equation}
\end{small} After calculating $\f{\beta^k {\B y^{(m+1)}}^T}{\beta {\B y^{(m)}}^k}$, we can further obtain the transformation matrix $\M_{m+1}$ from Eq.~(\ref{chain matrix beta}) and the unmixed partial derivative vector $\B v_m$ from Eq~.(\ref{fc vector}). The only left question is how to calculate the value of $\sigma^{(k)}(x)$ with $x\in \mathbb{R}$. Here we give the calculation of several widely used  activation functions.

\begin{itemize}
[itemsep=6pt,leftmargin=10pt]
    \item \textbf{Sine}
$\bm{\B \sigma (x)=sin(x).}$\quad
The derivatives are
\begin{equation}
\begin{split}
    \sigma^{(k)}(x)=
    \left\{
    \begin{array}{lc}
    cos(x), & k ~mod~ 4=1 \\
    -sin(x), & k ~mod~ 4=2 \\
    -cos(x), & k ~mod~ 4=3 \\
    sin(x). & k ~mod~ 4=0 \\
    \end{array}
    \right.
\end{split}
\end{equation}
\item  \textbf{ReLU}
$\bm{\sigma (x)=max(0,x)}$.\quad The derivatives are
\begin{equation}
\begin{split}
    \frac{\partial^k \sigma (x)}{\partial x^k}=
    \left\{
    \begin{array}{lc}
    1, & if~ k=1 ~and~ x>0 \\
    0. & else \qquad \qquad \qquad
    \end{array}
    \right.
\end{split}
\end{equation}

\item \textbf{Sigmoid}
$\bm{\sigma (x)=\frac{e^x}{1+e^x}}$.\quad The first derivative is 
\begin{equation}
    \sigma^{(1)} (x)
    =\frac{e^x}{(1+e^x)^2}
    =\sigma (x)-\sigma (x)^2.
\end{equation}
Further, we can express $\sigma^{(k)} (x)$ as the form containing only $\sigma (x)$, i.e.,
\begin{equation}
\begin{split}
    \sigma^{(2)} (x)&=\sigma^{(1)} (x)-2\sigma (x)\sigma^{(1)} (x) \\
    &=[\sigma (x)-\sigma (x)^2]-2\sigma (x)[\sigma (x)-\sigma (x)^2] \\
    &=\sigma (x)-3\sigma (x)^2+2\sigma (x)^3.
\end{split}
\end{equation}
After calculating the other derivatives and organizing them into matrix form, we have
\begin{equation}
\begin{split}
\left[
    \begin{array}{c}
            \sigma (x) \\
            \sigma^{(1)} (x) \\
            \sigma^{(2)} (x) \\
            \vdots \\
            \sigma^{(n)} (x)
    \end{array}
\right]=
\left[
    \begin{array}{cccc}
            1 & 0 & 0 & 0 \\
            1 & -1 & 0 & 0 \\
            1 & -3 & 2 & 0 \\
            \vdots & \vdots & \vdots & \ddots
    \end{array}
\right]
\left[
    \begin{array}{c}
            \sigma (x)  \\
            \sigma (x)^2  \\
            \sigma (x)^3  \\
            \vdots \\
            \sigma (x)^{n+1}
    \end{array}
\right].
\end{split}
\end{equation}
This square matrix takes a lower
triangular form, and we abbreviate it as $B\in \mathbb{R}^{n+1\times n+1}$. 
Similar to the derivation of the transformation
matrix in Eq.~(\ref{chain matrix}), from above equation, the $k$-th and $(k+1)$-th derivatives are
\begin{equation}
\begin{split}
    \sigma^{(k)} (x)=\sum_{i=1}^{k+1} B_{k+1,i}\sigma (x)^i,
\end{split}
\label{sigmoid formula 1}
\end{equation}
\begin{equation}
\begin{split}
    \sigma^{(k+1)} (x)=\sum_{i=1}^{k+2} B_{k+2,i}\sigma (x)^i.
\end{split}
\label{sigmoid formula 2}
\end{equation}

Taking derivatives over both sides of Eq.~(\ref{sigmoid formula 1}) yields
\begin{equation}
\begin{split}
    \sigma^{(k+1)} &(x)=\sum_{i=1}^{k+1} iB_{k+1,i}\sigma (x)^{i-1}\sigma^{(1)} (x) \\
    &=\sum_{i=1}^{k+1} iB_{k+1,i}\sigma (x)^{i-1}[\sigma (x)-\sigma (x)^2] \\
    &=\sum_{i=1}^{k+1} iB_{k+1,i}\sigma (x)^i-\sum_{i=1}^{k+1} iB_{k+1,i}\sigma (x)^{i+1} \\
    &=\sum_{i=1}^{k+1} iB_{k+1,i}\sigma (x)^i-\sum_{i=2}^{k+2} (i-1)B_{k+1,i-1}\sigma (x)^{i} \\
    &=\sum_{i=1}^{k+2}[iB_{k+1,i}-(i-1)B_{k+1,i-1}]\sigma (x)^i
\end{split}
\label{sigmoid formula 3}
\end{equation}
Comparing Eq.~(\ref{sigmoid formula 2}) and Eq.~(\ref{sigmoid formula 3}), we arrive at the following relationship
\begin{equation}  
\left\{  
    \begin{array}{lc}
        B_{1,1}=1, &  \\
        B_{i,j}=0, & i<j\\
        B_{i,j}=jB_{i-1,j}-(j-1)B_{i-1,j-1}. & i\geq j \\
    \end{array}  
\right.
\label{recurrence2}
\end{equation} 

\item \textbf{Tanh}
$\bm{\sigma (x)=\frac{e^x-e^{-x}}{e^x+e^{-x}}}$. \quad The  first and second derivatives are
\begin{equation}
\begin{split}
    &\sigma^{(1)} (x)=1-(\frac{e^x-e^{-x}}{e^x+e^{-x}})^2=1-\sigma (x)^2, \\
    &\sigma^{(2)} (x)=-2\sigma (x)\sigma^{(1)} (x)=-2\sigma (x)+2\sigma (x)^3.
\end{split}
\end{equation}

Organize it into matrix form:
\begin{equation}
\begin{split}
\left[
    \begin{array}{c}
            1 \\
            \sigma (x) \\
            \sigma^{(1)} (x) \\
            \sigma^{(2)} (x) \\
            \vdots \\
            \sigma^{(n)} (x)
    \end{array}
\right]=
\left[
    \begin{array}{ccccc}
            1 & 0 & 0 & 0 & 0 \\
            0 & 1 & 0 & 0 & 0 \\
            1 & 0 & -1 & 0 & 0 \\
            0 & -2 & 0 & 2 & 0 \\
            \vdots & \vdots & \vdots & \vdots & \ddots
    \end{array}
\right]
\left[
    \begin{array}{c}
            1 \\
            \sigma (x)  \\
            \sigma (x)^2  \\
            \sigma (x)^3  \\
            \vdots \\
            \sigma (x)^{n+1}
    \end{array}
\right].
\end{split}
\end{equation}
This square matrix takes a lower triangular form, and we abbreviate it as $\mathbf{C}\in \mathbb{R}^{n+2\times n+2}$. From above equation, the
$k$-th and $(k+1)$-th derivatives are
\begin{equation}
\begin{split}
    \sigma^{(k)} (x)=\sum_{i=1}^{k+2} C_{k+2,i}\sigma (x)^{i-1}, 
\end{split}
\label{tanh formula 1}
\end{equation}

\begin{equation}
\begin{split}
    \sigma^{(k+1)} (x)=\sum_{i=1}^{k+3} C_{k+3,i}\sigma (x)^{i-1}.
\end{split}
\label{tanh formula 2}
\end{equation}

Taking derivatives over both sides of Eq.~(\ref{tanh formula 1})
\begin{equation}
\begin{split}
    &\sigma^{(k+1)} (x)=\sum_{i=1}^{k+2} (i-1)C_{k+2,i}\sigma (x)^{i-2}\sigma^{(1)} (x) \\
    &=\sum_{i=1}^{k+2} (i-1)C_{k+2,i}\sigma (x)^{i-2}[1-\sigma (x)^2] \\
    &=\sum_{i=1}^{k+2} (i-1)C_{k+2,i}\sigma (x)^{i-2}-\sum_{i=1}^{k+2} (i-1)C_{k+2,i}\sigma (x)^{i} \\
    &=\sum_{i=0}^{k+1} iC_{k+2,i+1}\sigma (x)^{i-1}-\sum_{i=2}^{k+3} (i-2)C_{k+2,i-1}\sigma (x)^{i-1} \\
    &=\sum_{i=1}^{k+3}[iC_{k+2,i+1}-(i-2)C_{k+2,i-1}]\sigma (x)^{i-1} \\
\end{split}
\label{tanh formula 3}
\end{equation}
Comparing Eq.~(\ref{tanh formula 2}) and Eq.~(\ref{tanh formula 3}), we have
\begin{equation}  
\left\{  
    \begin{array}{lc}
        C_{1,1}=1, C_{2,1}=0, C_{2,2}=1, & \\
        C_{i,j}=0, & i<j \qquad \qquad  \\
        C_{i,j}=jC_{i-1,j+1}-(j-2)C_{i-1,j-1}. & i\geq j,i\geq 3
    \end{array}  
\right.
\label{recurrence3}
\end{equation} 

\end{itemize}
\vspace{2mm}

\subsection{Pooling Layer}
\begin{figure}[ht]
\begin{center}
\centerline{\includegraphics[width=0.92\columnwidth]{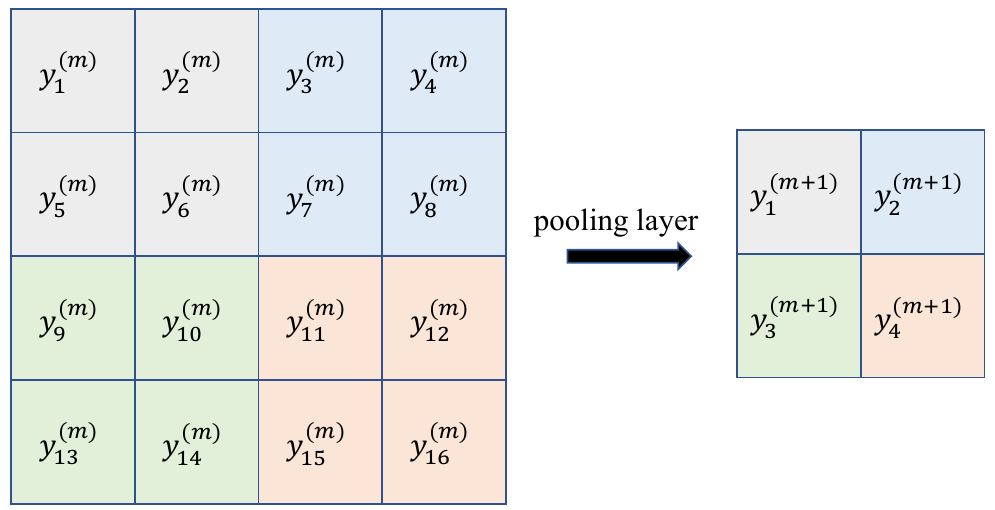}}
\caption{\textbf{The illustration of a simple 2-D pooling layer downsampling ${\mathbf y}^{(m+1)}$ to ${\mathbf y}^{(m)}$.} Here the kernel size is $2\times 2$, and the stride is $2\times 2$.}
\label{pooling layer}
\end{center}
\end{figure}

The pooling layer divides the input into blocks and takes either the maximal or average value of each block as the output. To demonstrate the expansion of the pooling layer, we take the simple 2-D pooling layer shown in Fig.~\ref{pooling layer} as an example.

\begin{figure}[t]
\begin{center}
\centerline{\includegraphics[width=0.92\columnwidth]{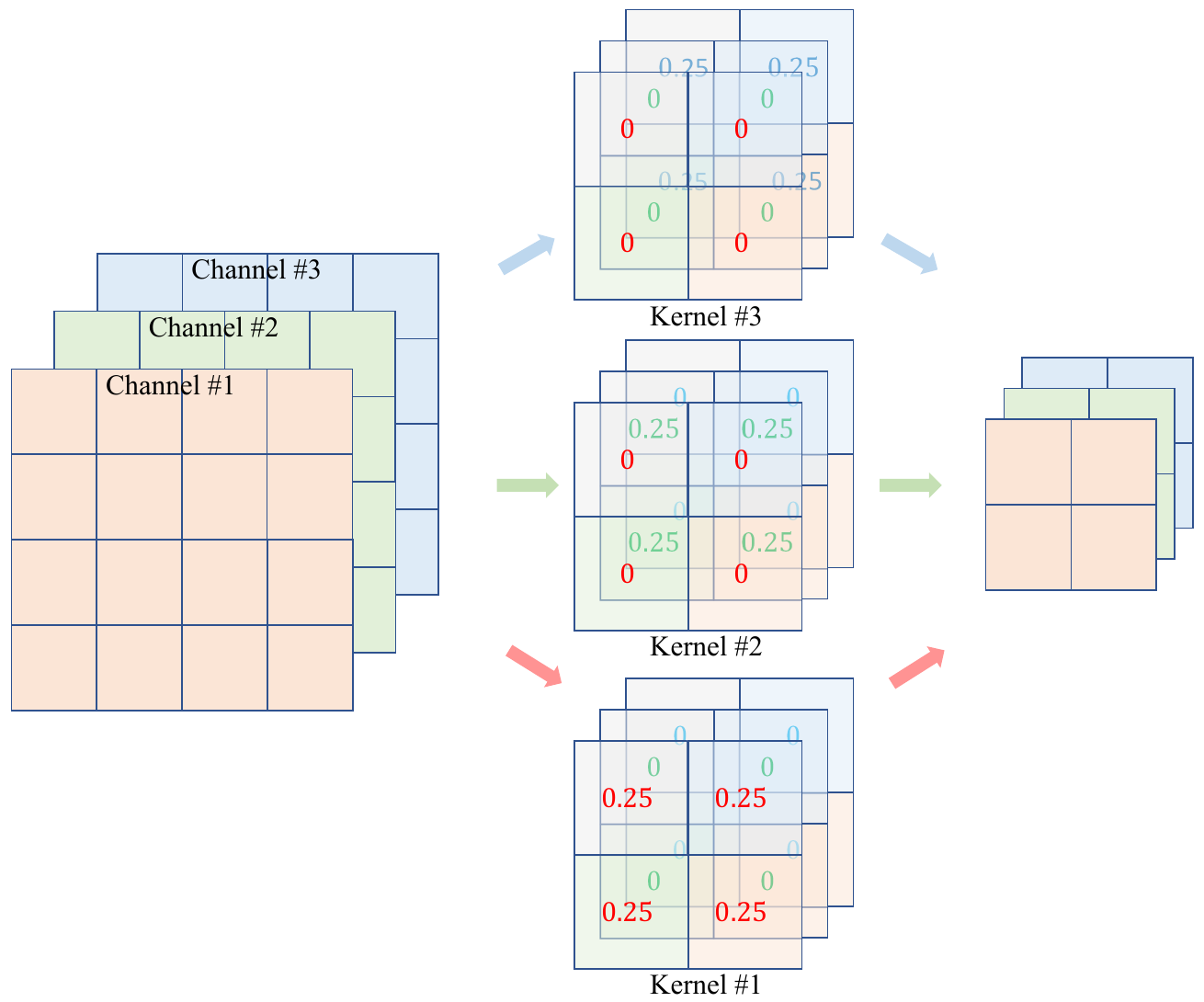}}
\caption{\textbf{The illustration of a convolutional layer equivalent to the average pooling layer.} Here the number of input and output channels are both 3, and 3 kernels are involved. 
Convolution the input with kernel $\#1$ 
acts as an average pooling applied specifically to the first input channel, and similar equivalence hold for the other two kernels.}
\label{average}
\end{center}
\end{figure}

\begin{itemize}
[itemsep=6pt,leftmargin=10pt]
    \item \textbf{Max pooling} layer outputs the maximum entry of each block, i.e.,
\begin{equation}  
\begin{split}
    \B y_1^{(m+1)}&=max(\B y_1^{(m)}, \B y_2^{(m)}, \B y_5^{(m)}, \B y_6^{(m)})=\B y^{(m)}_{idx_1},\\
    \B y_2^{(m+1)}&=max(\B y_3^{(m)}, \B y_4^{(m)}, \B y_7^{(m)}, \B y_8^{(m)})=\B y^{(m)}_{idx_2}, \\
    \B y_3^{(m+1)}&=max(\B y_9^{(m)}, \B y^{(m)}_{10}, \B y^{(m)}_{13}, \B y^{(m)}_{14})=\B y^{(m)}_{idx_3}, \\
    \B y_4^{(m+1)}&=max(\B y^{(m)}_{11}, \B y^{(m)}_{12}, \B y^{(m)}_{15}, \B y^{(m)}_{16})=\B y^{(m)}_{idx_4},
\end{split}
\end{equation} 
where $idx_1\in \{1,2,5,6\}, idx_2\in \{3,4,7,8\}, idx_3\in \{9,10,13,14\}, idx_4\in \{11,12,15,16\}$ are the subscripts of the maximum inputs in four blocks, respectively. Given $\f{\p \B y^k}{\p {\B y_i^{(m+1)}}^k}$, we can calculate the derivatives with respect to $\B y^{(m)}$ as
\begin{equation}  
\begin{split}
    \f{\p \B y^k}{\p {\B y^{(m)}_j}^k}=\left\{
    \begin{array}{cc}
       \f{\p \B y^k}{\p {\B y_i^{(m+1)}}^k},  & if~j=idx_i\\
       0. & else
    \end{array}
    \right.
\end{split}
\end{equation} 
Therefore, we only need to record the subscripts of the corresponding maximum inputs and assign $\f{\p \B y^k}{\p {\B y_i^{(m+1)}}^k}$ to $\f{\p \B y^k}{\p {\B y_{idx_i}^{(m)}}^k}$. When the stride is less than the kernel size, one input may be related to multiple outputs and the derivative can be written as the following general formula 
\begin{equation}  
\begin{split}
    \f{\p \B y^k}{\p {\B y^{(m)}_j}^k}=\sum_{\substack{i=1 \\ j=idx_i}}^{o_{m+1}} \f{\p \B y^k}{\p {\B y_i^{(m+1)}}^k}.
\end{split}
\end{equation} 

\item \textbf{Average pooling} layer takes the average over each block as the output, i.e.,
\begin{equation}  
\begin{split}
    \B y_1^{(m+1)}&=\f{1}{4}(\B y_1^{(m)}+\B y_2^{(m)}+\B y_5^{(m)}+\B y_6^{(m)}),\\
    \B y_2^{(m+1)}&=\f{1}{4}(\B y_3^{(m)}+\B y_4^{(m)}+\B y_7^{(m)}+\B y_8^{(m)}), \\
    \B y_3^{(m+1)}&=\f{1}{4}(\B y_9^{(m)}+\B y^{(m)}_{10}+\B y^{(m)}_{13}+\B y^{(m)}_{14}), \\
    \B y_4^{(m+1)}&=\f{1}{4}(\B y^{(m)}_{11}+\B y^{(m)}_{12}+\B y^{(m)}_{15}+\B y^{(m)}_{16}).
\end{split}
\end{equation} 
As shown in Fig.~(\ref{average}), we can regard a average pooling layer as a special convolutional layer, with the input channel being the same as the output channel. 
After getting the equivalent convolutional layer of the average pooling layer, we can obtain the derivatives easily from the high-order derivative rule of the convolutional layer.
\end{itemize}

\section{Taylor Expansion of Neural Networks}
\label{Taylor Expansion for Neural Networks}

As shown in Fig.~{\ref{propagation}}, applying a forward propagation and back-propagation, we can get a mixed partial derivative vector $\B v^\star_0$ on a reference input $\B x0$, which contains all the $n$-order derivatives $\{\f{\p^k\B y}{\p\B x_{i_1}\ldots\p\B x_{i_k}},i_1,\ldots,i_k=1,\ldots,p\}$. 

Denote the neural network as $\B y=f(\B x;\theta)$, with $\theta$ being the network parameters, the $n$-order Taylor polynomial of $f(\B x;\theta)$ can be expressed as 
\begin{equation}
\begin{split}
    f(\B x)&=f(\B x0;\theta) + \sum_{i=1}^{p} \f{\f{\p f(\B x;\theta)}{\p\B x_i}|_{\B x0}}{1!} \Delta\B x_i+\ldots \\
    &+\sum_{i_1,\ldots,i_n=1}^{p} \f{\f{\p^n f(\B x;\theta)}{\p\B x_{i_1}\ldots\p\B x_{i_n}}|_{\B x0}}{n!} \Delta\B x_{i_1}\ldots\Delta\B x_{i_n},
    \label{Taylor poly}
\end{split}
\end{equation}
where $\Delta\B x=\B x-\B x0$, $\f{\p^k f(\B x;\theta)}{\p\B x_{i_1}\ldots\p\B x_{i_k}}|_{\B x0}$ is a $k$-order partial derivative on the reference input $\B x0$.  
When all the modules are infinitely differentiable, the network is equivalent to an infinite Taylor polynomial. If the high-order derivatives are much smaller than the low-order derivatives, we can ignore the high-order terms and Eq.~(\ref{Taylor poly}) can provide an accurate and explicit explanation for the prediction.

\subsection{Upper and Lower Bounds of Network and Taylor Polynomial}
\label{Upper and Lower Bounds}
For simplicity, we take the 1-D neural network as an example (i.e., $\B x, \B x0 \in \mathbb{R}$). Suppose $f(\B x;\theta)$ has $n$-order continuous derivatives in the interval $[a, b]$ and $\B x0 \in (a,b)$, for $\forall \B x \in [a,b]$, $\exists \xi \in [\min(\B x, \B x0), \max(\B x, \B x0)]$, $s.t.$
\begin{equation}
f(\B x;\theta)=\sum_{k=0}^{n-1}\f{f(\B x0;\theta)^{(k)}}{k!}\Delta\B x^k + \f{f(\B \xi;\theta)^{(n)}}{n!}\Delta\B x^n,
\end{equation}
where $f(\B x;\theta)^{(k)}$ is the $k$-order derivative with respect to the input $\B x$, and $\f{f(\B \xi;\theta)^{(n)}}{n!}\Delta\B x^n$ is a $n$-order Lagrange remainder. 
After applying the proposed high-order Taylor expansion at $\B x0$, the $f(\B x;\theta)$'s $n$-order Taylor polynomial is derived as
\begin{equation}
f(\B x)=\sum_{k=0}^{n-1}\f{f(\B x0;\theta)^{(k)}}{k!}\Delta\B x^k + \f{f(\B x0;\theta)^{(n)}}{n!}\Delta\B x^n.
\end{equation}

Setting
\begin{equation}
\begin{split}
f_1(\B x)&=\sum_{k=0}^{n-1}\f{f(\B x0;\theta)^{(k)}}{k!}\Delta\B x^k + \f{\max_{\B x\in [a,b]} f(\B x;\theta)^{(n)}}{n!}\Delta\B x^n,\\
f_2(\B x)&=\sum_{k=0}^{n-1}\f{f(\B x0;\theta)^{(k)}}{k!}\Delta\B x^k + \f{\min_{\B x\in [a,b]} f(\B x;\theta)^{(n)}}{n!}\Delta\B x^n,
\label{function boundary0}
\end{split}
\end{equation}
we can provide the upper and lower boundaries of the network and the Taylor polynomial as
\begin{equation}
\begin{split}
    &f_d(\B x) \leq f(\B x;\theta) \leq f_u(\B x), \\
    &f_d(\B x) \leq f(\B x) \leq f_u(\B x),
\label{function boundary}
\end{split}
\end{equation}
where
\begin{equation}
\begin{split}
    f_u(\B x)&=\max\left(f_1(\B x),f_2(\B x)\right), \\
    f_d(\B x)&=\min\left(f_1(\B x),f_2(\B x)\right). \\
\end{split}
\end{equation}
Further, we can provide an upper boundary of the approximation error as
\begin{small}
\begin{equation}
\begin{split}
    |f(\B x)&-f(\B x;\theta)| \leq |f_u(\B x)-f_d(\B x)| 
    = |f_1(\B x)-f_2(\B x)|\\
    &= \f{\max_{\B x\in [a,b]}f(\B x;\theta)^{(n)}- \min_{\B x\in [a,b]}f(\B x;\theta)^{(n)}}{n!}|\Delta\B x|^n. \\
\end{split}
\label{error boundary}
\end{equation}
\end{small}

From the above equation, we can see that the approximation performance is closely related to three factors.
\begin{enumerate}
    \item The range of $f(\B x;\theta)^{(n)}$ in the interval $[a,b]$: $r=\max_{\B x\in [a,b]}f(\B x;\theta)^{(n)}-\min_{\B x\in [a,b]}f(\B x;\theta)^{(n)}$. Specifically, $r$ decreases with $f(\B x;\theta)^{(n+1)}$, and a smaller $r$ leads to a smaller approximation error. 
    \item The order of Taylor polynomial: $n$. When $n$ is large, the growth rate of $n!$ far exceeds the growth speed of $|\Delta\B x|^n$, resulting in a decrease in $|f(\B x)-f(\B x;\theta)|$.
    \item The distance from $\B x$ to the reference point $\B x0$: $|\Delta\B x|$. The closer $\B x$ is to $\B x0$, the smaller the approximation error tends to be.
\end{enumerate}

It should be noted that Eqns.~(\ref{function boundary})(\ref{error boundary}) are theoretical bounds for the error between neural networks and Taylor polynomial, and they hold only when the first $n$ derivatives are accurate enough. However, in practice, due to the precision limitations of computer storage and computation, it cannot be guaranteed that the upper bounds for the approximation error can always be estimated in all cases.

\subsection{Convergence Analysis of \name}
\label{Convergence}

If $\exists n$, $s.t.~\forall \B x \in [a,b]$, $|f(\B x;\theta)^{(n+1)}| \to 0$, we have 
$$\max_{\B x\in [a,b]}f(\B x;\theta)^{(n)}-\min_{\B x\in [a,b]}f(\B x;\theta)^{(n)} \to 0.$$
Further from Eq.~(\ref{error boundary}), the approximation error $|f(\B x)-f(\B x;\theta)| \to 0$ and then the Taylor polynomial will converge to the target neural network, i.e., $f(\mathbf{x}) \to f(\mathbf{x};\theta)$. In this section, we will analyze the condition of $|f(\B x;\theta)^{(n+1)}| \to 0$.

From Eqns.~(\ref{diag unmixed})(\ref{diag mixed})(\ref{conv k-order}), we can see that the $k$-order derivatives are related with $\B W_{i,j}^k$. 
\begin{equation}
\begin{split}
    |f(\B x;\theta)^{(k)}|\propto |\mathbf{W}_{i,j}|^k.
\end{split}
\end{equation}
When the elements in $\B W$ are concentrated near 0, high-order derivatives are more likely to approach 0. When the parameters are located far from 0, high-order derivatives may become increasingly larger, and thus the Taylor polynomial diverges, i.e., $f(\mathbf{x}) \nrightarrow f(\mathbf{x};\theta)$.
\begin{equation}
\begin{split}    
&\lim_{\substack{|\mathbf{W}_{i,j}| \to 0 \\ k \to \infty}} |f(\B x;\theta)^{(k)}| = 0,\\
&\lim_{\substack{|\mathbf{W}_{i,j}| >1 \\ k \to \infty}} |f(\B x;\theta)^{(k)}| = +\infty.
\end{split}
\end{equation}

The above analysis tells that the parameter distribution of each layer has a great influence on the convergence of Taylor expansion. Therefore, we can refer to the above rules to design network structures or impose constraints on the network parameters during network training to achieve deep neural networks with high-order Taylor approximation, which facilitates leveraging the advantages of such explicit expansion. We will verify this conclusion in the Experiment section.

\subsection{Time Complexity Analysis of \name}
As a fundamental building block and one of the most time-consuming operations in deep learning\cite{rumelhart1986learning}, back-propagation is implemented via automatic differentiation on the computational graph of the neural network in most deep learning frameworks, such as Autograd\cite{paszke2017automatic}. Here, we analyze and compare the time complexity of the computational-graph-based method and \name~ for a $p$-D neural network. 

When calculating high-order derivatives, the length of the computational graph increases exponentially with the order of the derivative at base 2, because for each node one needs to accumulate the local derivatives along all the paths from the node to the input, resulting in an exponential increase in the number of nodes in the computational graph. 
For computational-graph-based method, mathematically, there are $p^k$ $k$-order derivatives, and the length of their computational graphs is $2^{k-1}$, so the time complexity of the computational graph is
\begin{equation}
    T(n)=\sum_{k=1}^{n}p^k2^{k-1} \sim \mathcal{O}((2p)^n).
\end{equation}

Differently, \name~obtains all the derivatives at one time, with the main calculations lying in calculating the transformation matrix $\B M$ and conducting back-propagation. 
Since $\B M$ is a lower triangular matrix and the block matrices in $k$-th row need $k$ operations, the complexity of calculating $\B M$ is $T(n)=\sum_{k=1}^{n}k^2=\f{n(n+1)(2n+1)}{6}\sim\mathcal{O}(n^3)$. For linear layers, $\B M$ turns into a diagonal matrix and the complexity reduces to $T(n)=\sum_{k=1}^{n}k=\f{n(n+1)}{2}\sim\mathcal{O}(n^2)$. For mixed partial derivatives in Eq.~(\ref{abbreviated chain matrix beta mixed}), $\B M^\star$ is a diagonal matrix and the size of $\B Q_k$ is $p^{k-1}$ times larger than $\B W$, and the complexity is about $T(n)=\sum_{k=1}^{n}p^{k-1}=\f{1-p^n}{1-p}\sim\mathcal{O}(p^n)$. Therefore, the complexity of \name~ is
\begin{equation}
    \mathcal{O}(n^2)<T(n)<\mathcal{O}(p^n).
\end{equation}

\section{Experiments}
\label{Experiments}

In this section, we quantitatively 
demonstrate \name's significant advantages in terms of accuracy, speed, and memory consumption. We further explored the influence of the target network's parameter distribution on the convergence of its Taylor series, which verified the conclusion in Section \ref{Convergence}. Besides, we also visualized the 
upper and lower bounds of a neural network and its Taylor polynomial at increasing order of polynomial terms. 
Finally, we conducted three experiments to show \name's applications, including function discovery, low-cost inference, and feature selection.


\begin{figure*}[!t]
    \centering    \centerline{\includegraphics[width=0.88\textwidth]{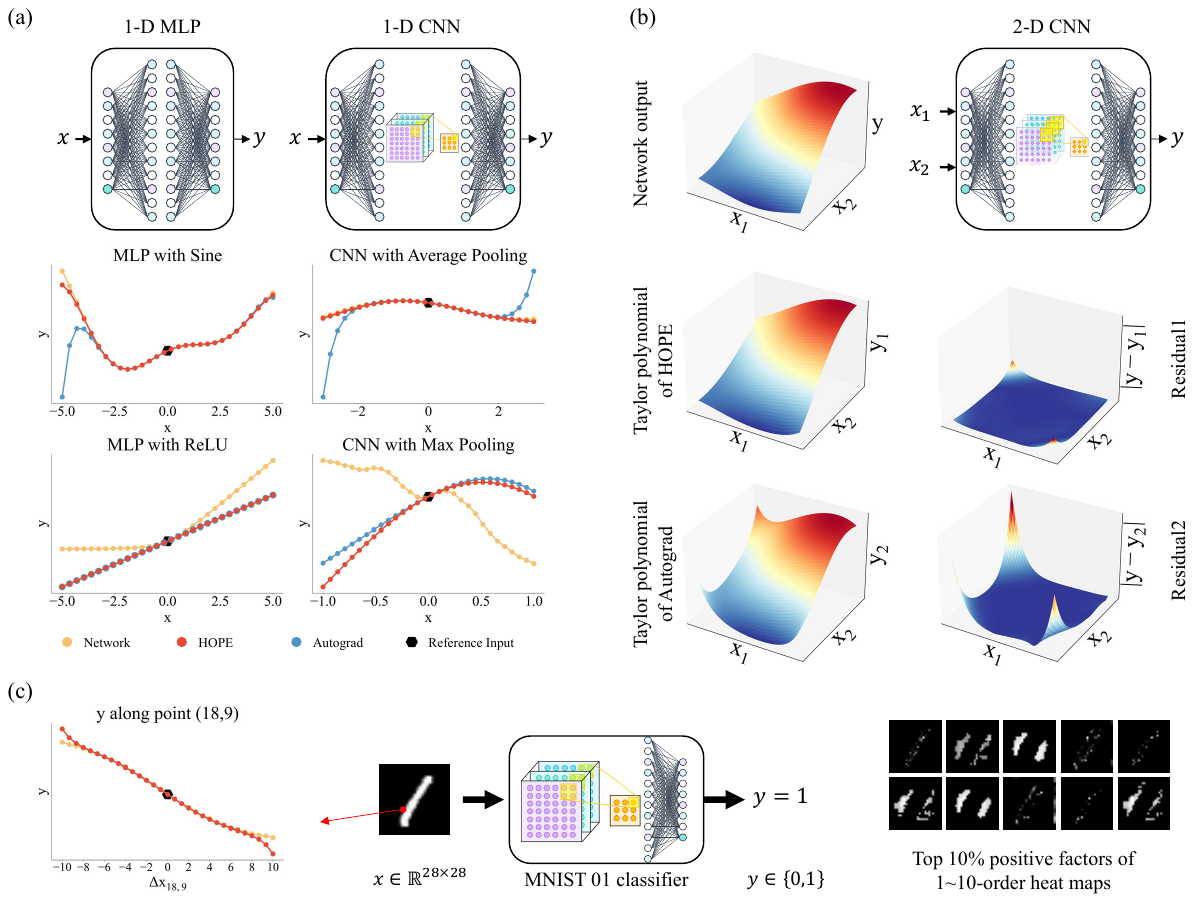}}
    \hfil
    \vspace{-4mm}
    \caption{\textbf{Approximation results of \name~ and Autograd.} (a) Approximation curves of 1-D networks with Sine (middle left), ReLU (bottom left), Average Pooling (middle right), and Max Pooling (bottom right). (b) Approximation surfaces (left) and approximation residuals (right) of a 2-D network. (c) Expansion results of an MNIST 01 classifier. High-order heat maps were calculated using \name~ (right), and a comparison (left) was made between the network and the Taylor polynomial with $\B x_{18,9}$ changed while keeping other input points unchanged.}
    \label{accuracy}
    \vspace{-3mm}
\end{figure*}

\subsection{Approximation Accuracy}
Among all the computational-graph-based methods, Autograd\cite{paszke2017automatic} stands out as the most widely used and convenient approach. Therefore, we have selected Autograd as the benchmark for comparison in this section.

Fig.~{\ref{accuracy}} shows the approximation accuracy of \name~ and Autograd on different neural networks. Specifically, we calculated the first 10-order derivatives with \name~ and Autograd separately, and get the Taylor polynomials with Eq.~(\ref{Taylor poly}). In Fig.~{\ref{accuracy}}(a), we compared the output curves of the 1-D network, the Taylor polynomial of \name, and Autograd. When all the modules of the network are $n$-order differentiable, \name~ can perform high local approximation on this neural network, while Autograd suffers from large deviation as the input moves far away from the reference point, which indicates that \name~ can get the high-order derivatives more accurately. When the network contains modules like ReLU and Max Pooling, both \name~ and Autograd can only obtain the first-order derivative.

In Fig.~{\ref{accuracy}}(b), we drew the input-to-output mapping surfaces and the residues of a 2D-CNN. One can observe that \name's output is closer to  the target surface, as shown in the residual of both methods, further demonstrating \name's superior accuracy in calculating high-order derivatives and capability to perform local approximations to neural networks. 

In Fig.~{\ref{accuracy}}(c), we trained an MNIST\cite{lecun1998gradient} 01 classifier, and get all the first 10-order unmixed partial derivatives on a certain input image with \name, while Autograd is unable to decompose this model because the input dimension is too large. We varied the intensity of point $\B x_{18,9}$ while keeping the other input points unchanged, and plot the input-to-output mapping of the network and the Taylor polynomial, showing the local approximation ability of \name. The top $10 \%$ positive factors of the first 10-order derivatives are shown on the right. Since previous works on network interpretability usually analyze the feature contribution from the first-order derivatives of the adopted neural network, while the visualized derivatives indicate that higher-order derivatives also reflect the influence of input on output, we will explore how to utilize higher-order heat maps in our future work.

\subsection{Running Efficiency}

We test the efficiency of the proposed approach \name~and Autograd, in terms of running time and memory costs. The experiments are conducted on the Windows operating system with a GPU of NVIDIA GeForce RTX 3080 Ti, 32 GB memory, and 20 CPU cores. The results are shown in Tab.~(\ref{time table}). 

We test on three MLPs with different input dimensions ($1\sim3$ ) but the same structure (10-layer MLP with a width of 1024), and an MNIST 01 classifier. We calculated all the mixed partial derivatives of the $1\sim3$ dimension networks, and for the MNIST network, we only calculate its unmixed partial derivatives. For a $p$-D network, there're $n~(p=1)$ or $\frac{p^{n+1}-p}{p-1}~(p>1)$ mixed partial derivatives, and for the MNIST network, there are $784n$ unmixed partial derivatives. 

\name~takes less than 1s and 6\% of the available memory in all cases, while the time consumption of Autograd is much longer and increases exponentially with $n$, or even out of memory (OOM). The significant superiority largely validates the time and memory efficiency of \name.

\begin{table*}[t]
\hfil
\vspace{-2mm}
\caption{\textbf{Comparison of time (s) / memory (\%) between \name~ and Autograd.} Here $n$ is the order of derivatives. When the program is not running, the system memory usage is recorded at 24\%. If an Out-of-Memory (OOM) error is encountered, it indicates that the program is utilizing 76\% or more of the available memory.}\label{time table}
\begin{center}
\vspace{-2mm}
\resizebox{\textwidth}{!}{
    \begin{scriptsize}
    \setlength{\tabcolsep}{0.5mm}{
        \begin{tabular}{c||cc|cc|cc|cc} 
            \hline
                &
                \multicolumn{2}{c|}{$\B x\in \mathbb{R}$} & 
                \multicolumn{2}{c|}{$\B x\in \mathbb{R}^2$} & 
                \multicolumn{2}{c|}{$\B x\in \mathbb{R}^3$} & 
                \multicolumn{2}{c}{$\B x\in \mathbb{R}^{28\times 28}$} \\
                \cline{2-9}
                $\B n$ 
                & \scriptsize \quad\quad\name\quad\quad~ & \scriptsize \quad\quad~ Autograd \quad\quad
                & \scriptsize \quad\quad\name\quad\quad~ & \scriptsize \quad\quad~ Autograd \quad\quad
                & \scriptsize \quad\quad\name\quad\quad~ & \scriptsize \quad\quad~ Autograd \quad\quad
                & \scriptsize \quad\quad\name\quad\quad~ & \scriptsize \quad\quad~ Autograd \quad\quad\\ 
            \hline
                1 & 0.0099 / 0.20 & 0.0328 / 0.10 & 0.0109 / 0.30 & 0.0617 / 0.10 & 0.0120 / 0.20 & 0.0462 / 0.10 & 0.0120 / 0.10 & 0.0339 / 0.80 \\
                2 & 0.0159 / 0.40 & 0.0358 / 0.10 & 0.0160 / 0.30 & 0.0820 / 0.10 & 0.0162 / 0.30 & 0.0921 / 0.20 & 0.0431 / 0.10 & 3.3456 / 2.10 \\
                3 & 0.0229 / 0.40 & 0.0438 / 0.10 & 0.0232 / 0.30 & 0.1278 / 0.10 & 0.0234 / 0.40 & 0.2558 / 0.20 & 0.0229 / 0.10 & 7.5919 / 4.80 \\
                4 & 0.0428 / 0.50 & 0.0538 / 0.10 & 0.0448 / 0.60 & 0.3559 / 0.20 & 0.0468 / 0.50 & 1.1251 / 0.90 & 0.0342 / 0.10 & 18.457 / 10.8 \\
                5 & 0.0649 / 0.60 & 0.0812 / 0.10 & 0.0669 / 0.60 & 1.1938 / 0.80 & 0.0711 / 0.60 & 6.9751 / 5.80 & 0.0399 / 0.10 & 101.02 / 25.7 \\
                6 & 0.0857 / 0.80 & 0.1446 / 0.10 & 0.0867 / 0.80 & 5.0142 / 4.00 & 0.0974 / 0.80 & 59.799 / 43.7 & 0.0479 / 0.10 & 1129.2 / 56.4 \\
                7 & 0.1086 / 0.80 & 0.3037 / 0.20 & 0.1196 / 0.90 & 31.890 / 21.2 & 0.1340 / 1.00 & OOM / 76+ & 0.0648 / 0.10 & OOM / 76+ \\
                8 & 0.1387 / 0.90 & 0.7244 / 0.50 & 0.1405 / 0.90 & OOM / 76+ & 0.1839 / 1.40 & OOM / 76+ & 0.0731 / 0.20 & OOM / 76+ \\
                9 & 0.1634 / 1.00 & 1.9066 / 1.50 & 0.1638 / 1.10 & OOM / 76+ & 0.3234 / 2.40 & OOM / 76+ & 0.0864 / 0.30 & OOM / 76+ \\
                10 & 0.1942 / 1.20 & 5.4099 / 4.10 & 0.1982 / 1.50 & OOM / 76+ & 0.7079 / 5.10 & OOM / 76+ & 0.1000 / 0.30 & OOM / 76+ \\      
    \bottomrule
    \end{tabular}
    }
    \end{scriptsize}
}
\end{center}
\vskip -0.1in
\end{table*}

\subsection{Convergence Under Different Parameter Settings}

Tab.~(\ref{converge table}) shows the influence of the parameter distribution of the target neural network on the convergence of its Taylor series. We initialize the weights of an MLP (width 512, depth 5) to follow a uniform distribution $\B W^{(m+1)}_{i,j}\sim U(-\frac{w_0}{o_{m}},\frac{w_0}{o_{m}})$. As the value of $w_0$ decreases, the parameters tend to concentrate more closely around zero, and it is more likely that the high-order derivatives of the model become increasingly smaller according to the inference in Section \ref{Convergence}. The data in this table is the absolute value of the $n$-order derivative divided by the first-order derivative (i.e., $|\frac{\partial^n f}{\partial x^n}/\frac{\partial f}{\partial x}|$). We varied $w_0$ from 0.01 to 100. When $w_0=0.01$ and $w_0=0.1$, the high-order derivatives are much smaller than the low-order derivatives. When $w_0=1$, almost all of the derivatives are on the same order of magnitude. When $w_0=10$ and $w_0=100$, the high-order derivatives are far larger than the low-order derivatives, which means that we cannot ignore the higher-order terms and make local approximations to neural networks. 

\subsection{Upper and Lower Bounds of a Neural Network and its Taylor Polynomial}

\begin{figure*}[t]
    \begin{center}   
    \vspace{-3mm}    \centerline{\includegraphics[width=0.96\textwidth]{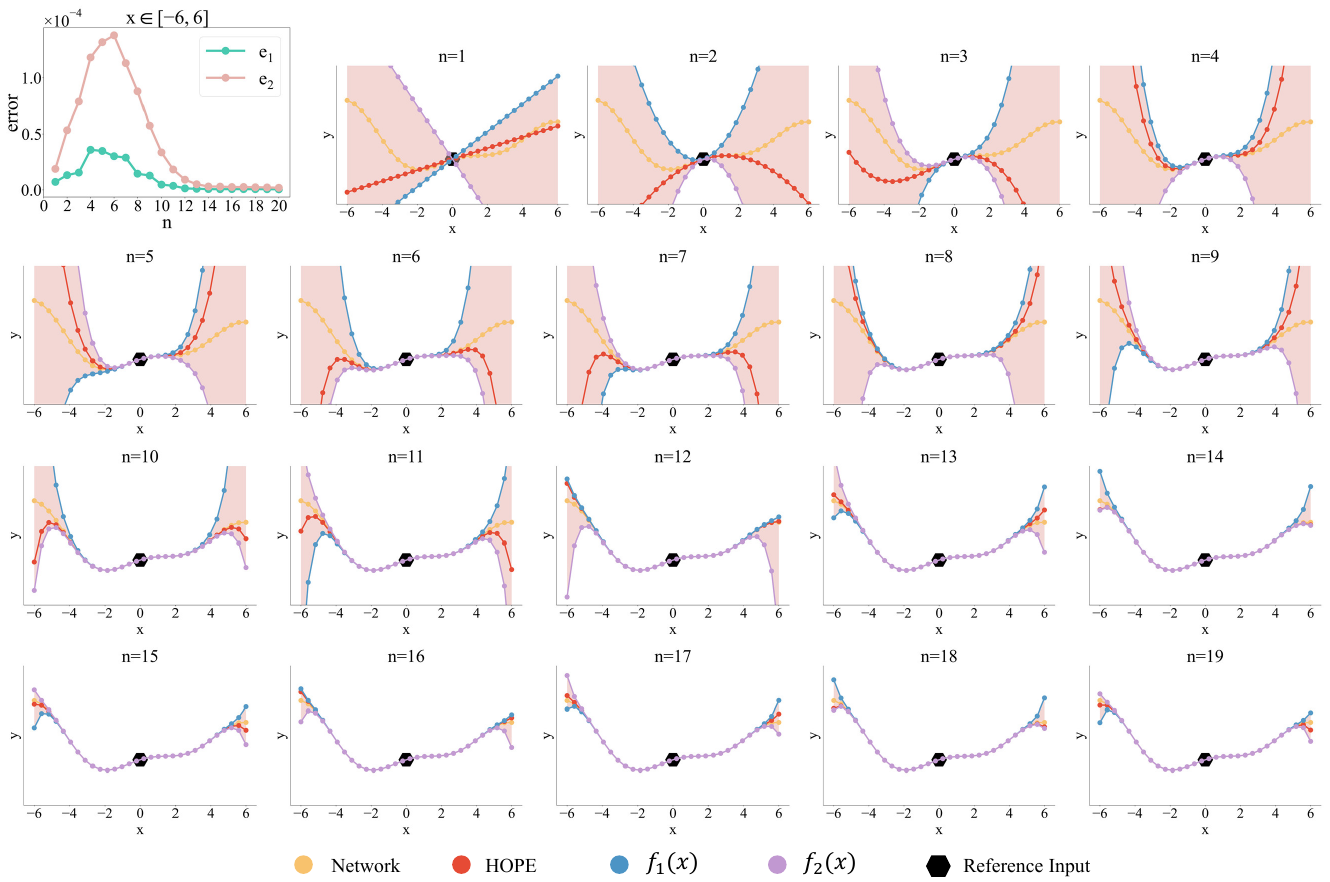}}
    \vspace{-2mm}
    \caption{\textbf{Upper and lower bounds of a network and its Taylor polynomials of different orders.} The first figure shows the errors between a neural network and the $n$-order Taylor polynomials in the interval $[-6,6]$. Here $e_1=\max_{\B x\in [-6,6]} |f(\B x)-f(\B x;\theta)|$, and  $e_2=\f{\max f(\B x;\theta)^{(n)}- \min f(\B x;\theta)^{(n)}}{n!}6^n$. 
    $f_1(x)$ and $f_2(x)$ are respectively the theoretical bound of the neural network and Taylor polynomial.
    } 
    \label{boundary}
    \end{center}
    \vspace{-3mm}
\end{figure*}

Based on Eq.~\ref{error boundary}, one can calculate the theoretical error bound between a neural network and its Taylor polynomial. In Fig.~\ref{boundary}, the first panel shows the maximum approximation error $e_1$ in the interval [-6,6] at different orders, and the theoretical upper bound of error $e_2$. A small $e_1$ or $e_2$ indicates that the model has small prediction errors at each point. One can see that the theoretical error $e_2$ is always larger than $e_1$, and when $n>14$ the magnitude of $e_2$ reduces to a small value, resulting in a high approximation accuracy (small $e_1$). 

\begin{table}[H]
\caption{\textbf{Convergence of the Taylor series.} The weights of each layer follow a uniform distribution $\B W^{(m+1)}_{i,j}\sim U(-\frac{w_0}{o_{m}},\frac{w_0}{o_{m}})$, with $w_0$ controlling the parameter distributions. The scores in each cell is the ratio between the absolute value of the $n$-order derivative and the first-order derivative (i.e., $|\frac{\partial^n f}{\partial x^n}/\frac{\partial f}{\partial x}|$), indicating the convergence.}
\label{converge table}
\begin{center}
\resizebox{\columnwidth}{!}{
    \begin{scriptsize}
    \setlength{\tabcolsep}{1.0mm}{
    \begin{tabular}{c||c|c|c|c|c} 
        \hline
            $n$ &~~$w_0$=0.01~~& ~~$w_0$=0.1~~& ~~$w_0$=1.0~~& ~~$w_0$=10~~& ~~$w_0$=100~~\\ 
        \hline
            1~ & 1.00e+00 & 1.00e+00 & 1.00e+00 & 1.00e+00 & 1.00e+00 \\
            2~ & 4.69e-03 & 6.39e-02 & 7.36e-02 & 3.14e+01 & 2.98e+03 \\
            3~ & 3.94e-05 & 6.99e-03 & 7.98e-01 & 3.83e+01 & 6.08e+03 \\
            4~ & 3.12e-07 & 5.18e-04 & 1.10e-01 & 2.28e+03 & 5.27e+07 \\
            5~ & 1.97e-09 & 5.30e-05 & 7.01e-01 & 3.49e+03 & 1.66e+08 \\
            6~ & 2.15e-11 & 4.42e-06 & 1.28e-01 & 2.15e+05 & 1.39e+12 \\
            7~ & 1.28e-13 & 4.04e-07 & 6.37e-01 & 1.49e+05 & 5.56e+12 \\
            8~ & 1.62e-15 & 3.76e-08 & 1.35e-01 & 3.04e+07 & 4.22e+16 \\
            9~ & 1.07e-17 & 2.94e-09 & 5.85e-01 & 2.49e+07 & 1.95e+17 \\
            10~ & 1.31e-19 & 3.20e-10 & 1.39e-01 & 7.30e+09 & 1.23e+21 \\
    \bottomrule
    \end{tabular}
    }
    \end{scriptsize}
    }
\end{center}
\vskip -0.1in
\end{table}

Other panels in Fig.~\ref{boundary} show the evolution of the approximation curves with increasing orders. $f_1(x)$ and $f_2(x)$ are the theoretical bounds of the network and its Taylor polynomial, and their expressions can be found in Eq.~\ref{function boundary0}. As the degree of the approximation increases, the bounds $f_1(x)$ and $f_2(x)$ gradually converge, resulting in a closer approximation between this neural network and its Taylor polynomial. 

\subsection{Applications} 

\noindent\textbf{Interpretation of deep neural networks and function discovery.\quad} Implicit neural functions \cite{mildenhall2021nerf,sitzmann2020implicit,tancik2020fourier,yang2022sci,yang2022sharing} are of strong expression capability and can be used to describe some unknown systems from observations. Taking the 2-D function 
\begin{equation}
    y=\f{\B x_1^2+\B x_2}{2}, ~\B x1,\B x2\in [-1,1]
    \label{f1}
\end{equation}
as an example, we uniformly sample in the range $[-1,1]^2$ and then use a 2-D MLP to fit its observations. 
Since the MLP is a ``black-box'', we expand it into 2-order Taylor polynomials on reference inputs (0.0,~0.0), (0.5,~0.5), and (-0.5,~-0.5) separately, and achieve following explicit expressions
\begin{small}
\begin{equation}
\begin{split}
\B y&=-0.01+0.00\B x_1+0.51\B x_2+0.55\B x_1^2-0.00\B x_1\B x_2+0.03\B x_2^2 \\
    &\approx 0.51\B x_2+0.55\B x_1^2, \\
\B y&=0.38+0.54(\B x_1-0.5)+0.51(\B x_2-0.5)+
    0.53(\B x_1-0.5)^2 \\
    &+0.06(\B x_1-0.5)(\B x_2-0.5)+0.03(\B x_2-0.5)^2 \\
    &=0.01-0.02\B x_1+0.45\B x_2+0.53\B x_1^2+0.06\B x_1\B x_2+0.03\B x_2^2 \\
    &\approx 0.45\B x_2+0.53\B x_1^2, \\
\B y&=-0.13-0.51(\B x_1+0.5)+0.49(\B x_2+0.5)+
    0.50(\B x_1+0.5)^2 \\
    &-0.07(\B x_1+0.5)(\B x_2+0.5)+0.02(\B x_2+0.5)^2 \\
    &=-0.03-0.05\B x_1+0.48\B x_2+0.50\B x_1^2-0.07\B x_1\B x_2+0.02\B x_2^2 \\
    &\approx 0.48\B x_2+0.50\B x_1^2.
\end{split}
\vspace{-3mm}
\end{equation}
\end{small}

The aforementioned equations provide local explanations for the ``black-box'' network. When all these local explanations align and reach a consistent conclusion, a global explanation can be obtained. The findings suggest that \name~ possesses the capability of local interpretation and also exhibits potential for global interpretation and function discovery. 
Furthermore, the results validate that the expanded polynomial can learn the latent function with the same fidelity as the trained neural network, 
within the whole interval $[-1,1]^2$. This also implies that 
\name~ can be employed to assess the quality of the model.

\begin{figure}[t]
    \begin{center}    \centerline{\includegraphics[width=0.5\textwidth]{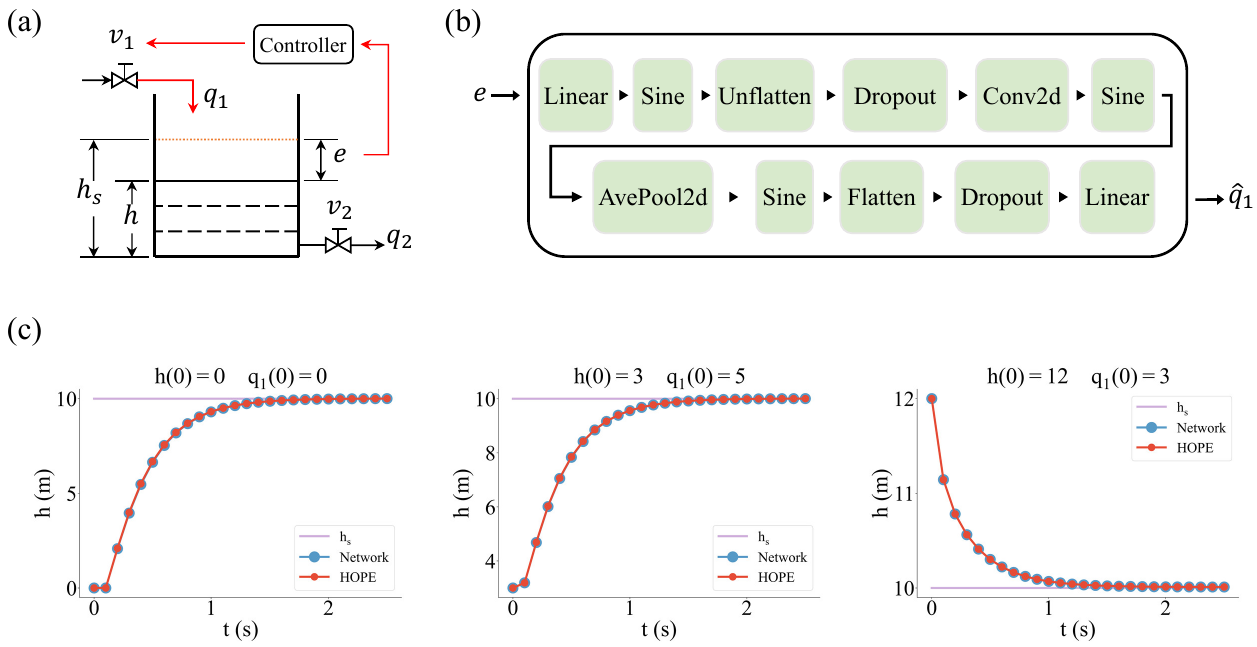}}
    \vspace{-2mm} \caption{\textbf{Illustration of the single-tank liquid level control system.} (a) A single-tank system. (b) The structure of the deep neural network implementing the controller, with the name of each module corresponds to the name of the class invoked in PyTorch. (c) Liquid level control curves of the neural-network controller and its polynomial expansion under different initial liquid levels and inflow rates.} 
    \label{tank}
    \end{center}
    \vspace{-5mm}
\end{figure}

\vspace{2mm}
\noindent\textbf{Low-cost inference of deep neural networks.\quad}
To show the advantageous running efficiency after Taylor expansion, we test on a controller of a single-tank liquid system implemented with a neural network. We simulated the following liquid system, in which the opening of the water outlet valve $v_2$ keeps constant, while the water output $q_2$ is determined by the liquid level height $h$, as illustrated in Fig.~\ref{tank}(a). Specifically, to achieve the desired liquid level height $h_s$, the opening of the inlet valve $v_1$ is manipulated to regulate the quantity of inlet water $q_1$, and the system is a first-order differential system, with transfer function
\begin{equation}
    G(s) = \f{K}{Ts+1}.
\end{equation}
Here $K$=1 is the system's gain, $T$=1 is the system's time constant, and $s$ is the complex frequency domain variable. 

We trained a neural network level controller $\hat{q_1}=f(e;\theta)$, where the input is the liquid level difference $e=h_s-h$, and the output is the estimated input flow rate $\hat{q_1}$. The network structure is shown in Fig.~\ref{tank}(b). Here we set $h_s=10$, and the label for training is designed as
\begin{equation}
\begin{split}
    f(e)=\left\{
    \begin{array}{cc}
        20, & 6<e\leq 10 \\
        15, & 2<e\leq 6 \\
        10, & -2<e\leq 2 \\
        5, & -6<e\leq -2 \\
        0. & -10<e\leq -6
    \end{array}
    \right.
\end{split}
\end{equation}
We normalize both the inputs and labels to the range of -1 to 1, and the loss function is defined as
\begin{equation}
    J(e;\theta)= \Vert f(\f{e}{10};\theta)- (\f{f(e)}{10}-1)\Vert_2.
\end{equation}
Due to the continuity of this neural network, $f(e;\theta)$ will be relatively smooth and not perfectly fit the function $f(e)$. 

We expand $f(e;\theta)$ to 3-order Taylor polynomials,  
within the range [-1, 1], to replace the neural network perfectly. 
We conducted liquid-level control experiments on the system using both the original neural network and the 3-order polynomial, under different initial liquid levels and initial inflow rates. The results are shown in Fig.~\ref{tank}(c), from which we can see that the polynomial can exactly replace the neural network.

\begin{table}[H]
\caption{\textbf{Inference time (ms) under different input batches.} 
}
\label{inference time}
\vspace{-3mm}
\begin{center}
\resizebox{\columnwidth}{!}{
    \begin{scriptsize}
    \setlength{\tabcolsep}{1.0mm}{
    \begin{tabular}{c||c|c|c|c|c|c|c} 
        \hline
            Batch size & 1 & 4 & 16 & 64 & 256 & 1024 & 4096 \\ 
        \hline
            Network & 0.2056 & 0.2640 & 0.4163 & 2.5155 & 10.805 & 36.401 & 121.98 \\
            HOPE & 0.0284 & 0.0284 & 0.0284 & 0.0284 & 0.0291 & 0.0294 & 0.0324\\
    \bottomrule
    \end{tabular}
    }
    \end{scriptsize}
    }
\end{center}
\vskip -0.1in
\end{table}

We conducted statistical analysis on the inference time of the deep neural network and its polynomial approximation by \name, across different input batch sizes. The results in Tab.~(\ref{inference time}) indicate that the expanded polynomial has significantly shorter and more concentrated inference time. When the input batch grows to 4096, the polynomial inference time is approximately 3765 times faster than that of the neural network, while maintaining the same prediction accuracy. 
In terms of the model size, the file size of the deep neural network reaches 44,671 bytes, while the Taylor polynomial merely occupies 160 bytes, showcasing a remarkable reduction in storage space.

\begin{figure}[t]
    \begin{center}    \centerline{\includegraphics[width=\linewidth]{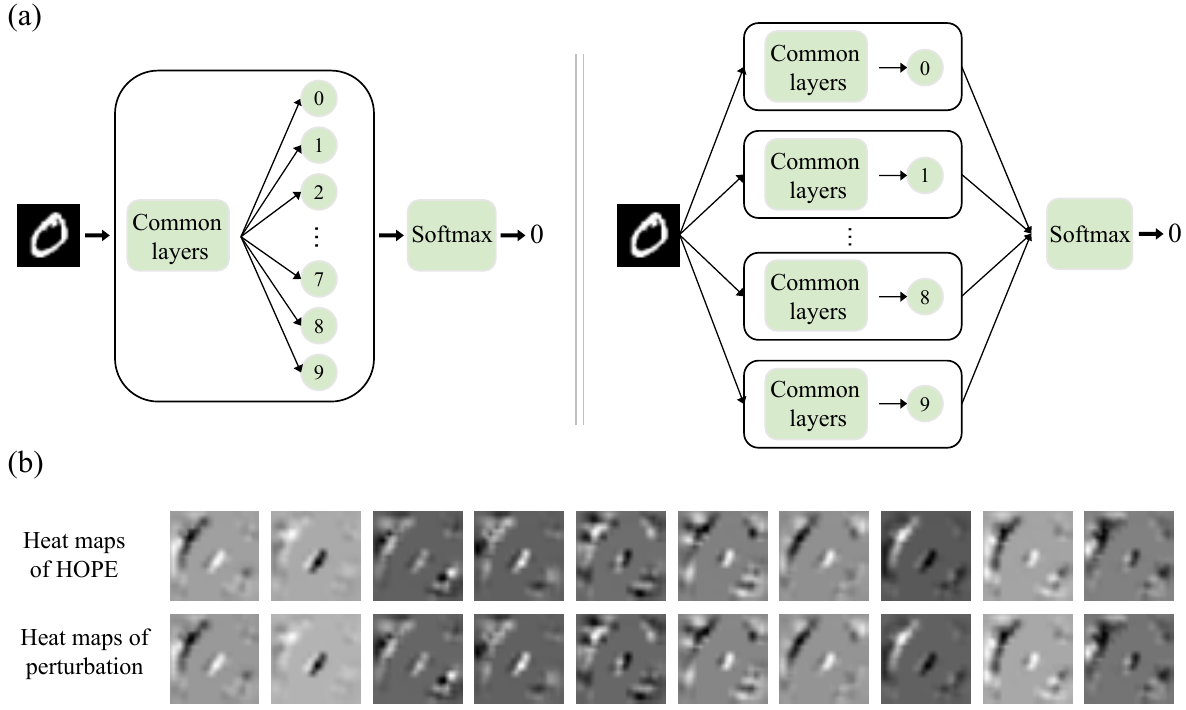}}
    \caption{\textbf{Heat maps of an MNIST digit classifier.} (a) The MNIST classifier and the ten separated single-output models. (b) Comparison of heat maps by \name~ and perturbation-based method.
    } 
    \label{heatmaps}
    \end{center}
    \vspace{-2mm}
\end{figure}

\vspace{2mm}
\noindent\textbf{Feature selection.\quad} 
For a neural network taking multiple inputs, an explicit expansion can efficiently measure the quantitative contribution of each input element to the final output, and facilitate feature selection. To demonstrate this application, we trained an MNIST handwritten digit classifier, and then separate it into ten equivalent single-output classifiers for easier expansion, 
as illustrated in Fig.~\ref{heatmaps}(a). 
Denoting the input image as $\B x$, prediction as $\B y$, and $\B x_{i,j}$'s impact on $\B y$ as $\Delta \B y_{i,j}\in \mathbb{R}$, we have 
\begin{equation}
\begin{split}
    \Delta \B y_{i,j}\approx \sum_{k=1}^n \f{\f{\p^k \B y}{\p \B x_{i,j}^k}}{k!}\Delta \B x_{i,j}^k,
\end{split}
\end{equation}
which can be further converted into matrix form
\begin{equation}
\begin{split}
    \Delta \B y\approx \sum_{k=1}^n \f{\f{\beta^k \B y}{\beta \B x^k}}{k!}\odot \Delta \B x^{\circ k}.
\end{split}
\end{equation}
Here $\Delta \B x\in \mathbb{R}^{28\times 28}$ represents the perturbation applied to the input image, $\f{\beta^k \B y}{\beta \B x^k}\in \mathbb{R}^{28\times 28}$ contains all the $k$-order unmixed partial derivatives, and $\Delta \B y\in \mathbb{R}^{28\times 28}$ is the heat map reflecting the impact of all input elements on the output.

We initialize $\Delta \B x$ as $\B 1^{28\times 28}$ and expanded the ten single-output models to obtain 10 heat maps, and then applied perturbation analysis to get another 10 counterparts, as visualized in the upper and lower rows of Fig.~\ref{heatmaps}(b). 
The results demonstrate that \name~ is capable of generating an equivalent heat map to the perturbation-based method. Besides, \name~can infer significantly faster, taking only 0.002s to generate the heat map, whereas perturbation analysis requires 0.342s.

\section{Conclusion}
\label{Conclusion}

Aiming at providing a high precision polynomial interpretation of the ``black-box'' deep neural networks, we propose the network's high-order Taylor expansion, which is of high accuracy, low computational cost, good convergence and wide applications. Understanding the mechanism behind the deep neural networks is quite important, and we believe that neural networks will become more transparent with \name, accelerating the development and application of neural networks.

\vspace{2mm}
\noindent{\bf Summary of the approach.\quad}
Specifically, we first derive the high-order derivative rule for a general composite function and then extend the rule to neural networks for fast and accurate calculation of its high-order derivatives. From all above derivatives, we can expand a black-box network into an explicit Taylor polynomial, providing a local explanation for the network's mapping from the input to the output. We also theoretically prove that a neural network is equivalent to its infinite Taylor polynomial if all of the modules are infinitely differential, and analyze the polynomial's convergence condition as well. 

\vspace{2mm}
\noindent{\bf Advantageous and applications.\quad}
\name~ has significant advantages in terms of accuracy, speed, and memory cost compared with computational-graph-based method. It works as a general expansion and thus of wide applicability for diverse deep neural networks, e.g., with different dimensions and layers.

The explicit Taylor expansion possesses the ability to conduct data-driven explicit model identification, facilitate the fast inference of a trained deep neural network, and select informative features contributing to the output.


\vspace{2mm}
\noindent{\bf Limitations and future work.\quad}
It should be noted that \name~ can be used only for modules that are $n$-order differentiable. For networks including components such as ReLU, LeakReLU, or Max Pooling, both \name~ and Autograd can only obtain their first-order information. Further increasing the applicability is one of our ongoing work. 

In the future, we would like to explore deeper the relationship between the convergence of the Taylor series and the parameter distribution, and apply it to the optimization and structure design of deep neural networks. We will also explore how to use high-order heat maps to  determine the contribution of inputs to outputs more accurately. Moreover, \name~ can get the derivatives between any nodes of a neural network, which might inspire lightweight network design. 

\section*{Acknowledgments}

This work is jointly funded by National Natural Science Foundation of China (Grant No. 61931012) and Beijing Municipal Natural Science Foundation (Grant No. Z200021).

\ifCLASSOPTIONcaptionsoff
  \newpage
\fi



%



\bibliographystyle{IEEEtran}
\bibliography{main}

%

\begin{IEEEbiography}[{\includegraphics[width=1in,height=1.25in,clip,keepaspectratio]{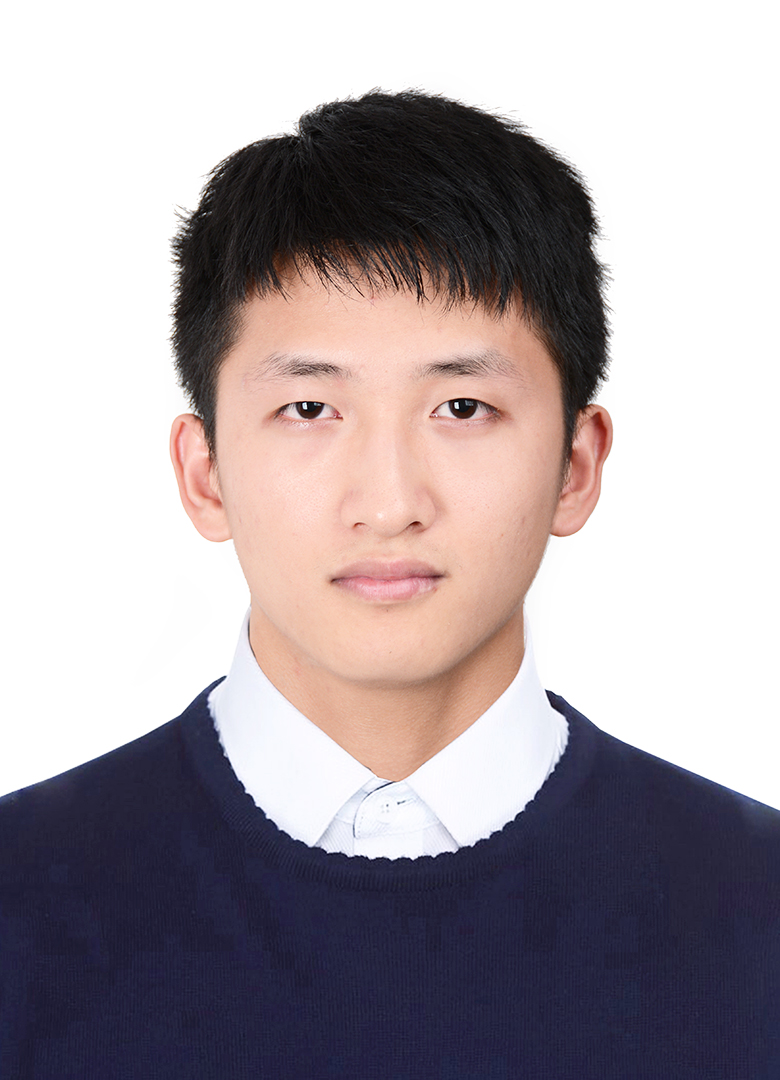}}]{Tingxiong Xiao} 
received the B.Eng. degree from the Department of Automation, Wuhan University,
Wuhan, China, in 2022. He is currently a Ph.D. student in the Department of Automation, Tsinghua University, Beijing, China. His research interests mainly include machine learning and computer vision.
\end{IEEEbiography}

\begin{IEEEbiography}[{\includegraphics[width=1in,height=1.25in,clip,keepaspectratio]{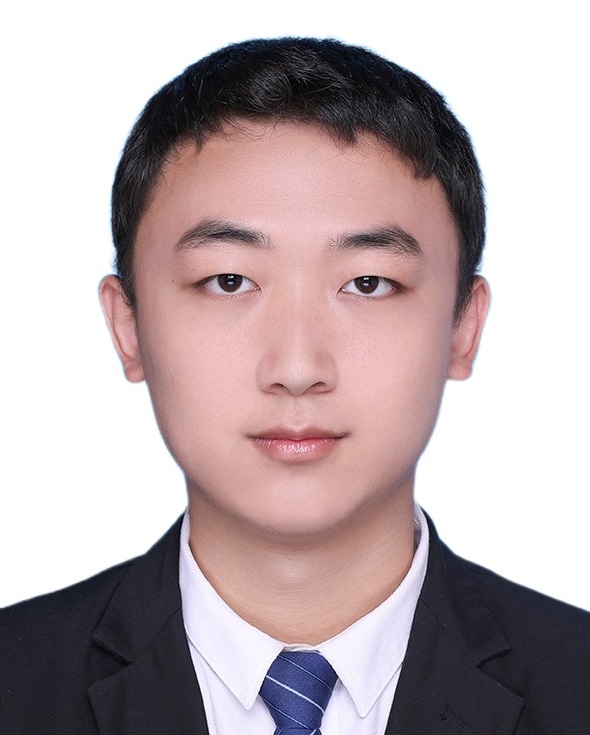}}]{Weihang Zhang}
received the B.S. degree in automation from Tsinghua University, Beijing, China, in 2019. He is currently working toward the Ph.D. degree in control theory and engineering in the Department of Automation, Tsinghua University, Beijing, China. His research interests include computational imaging, microscopy, and computer vision.
\end{IEEEbiography}

\begin{IEEEbiography}[{\includegraphics[width=1in,height=1.25in,clip,keepaspectratio]{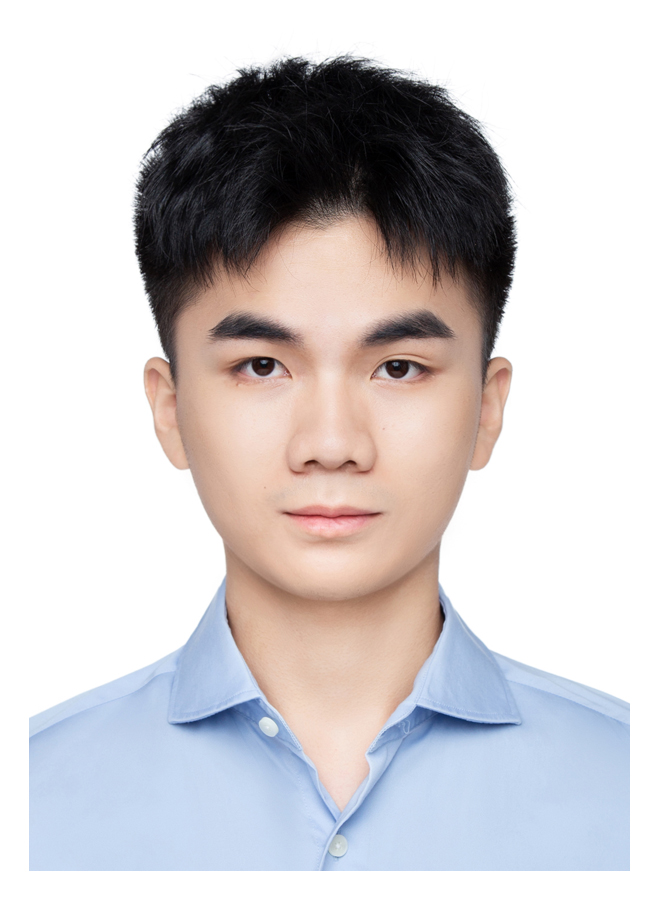}}]{Yuxiao Cheng}
received the B.Eng. degree from the Department of Automation, Tsinghua University, Beijing, China, in 2022. He is currently a Ph.D. student in the Department of Automation, Tsinghua University. His research interest is machine learning.
\end{IEEEbiography}

\begin{IEEEbiography}[{\includegraphics[width=1in,height=1.25in,clip,keepaspectratio]{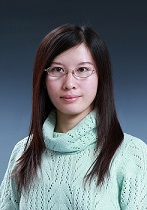}}]{Jinli Suo}
received the B.S. degree in computer science from Shandong University, Jinan, China, in 2004 and the Ph.D. degree in computer application technology from the Graduate University of Chinese Academy of Sciences, Beijing, China, in 2010. She is currently a tenured associate professor with the Department of Automation, Tsinghua University, Beijing, China. Her research interests include computer vision, computational photography, and statistical learning. She serves as an Associate Editor for IEEE Transactions on Computational Imaging and the Journal of the Optical Society of America A.
\end{IEEEbiography}

\vfill

\end{document}